%% file: main.tex
\pdfoutput=1
\documentclass{article}
\usepackage{preprint,times}

\usepackage[utf8]{inputenc}
\usepackage[T1]{fontenc}
\usepackage{hyperref}
\usepackage{url}
\usepackage{graphicx}
\usepackage{booktabs}
\usepackage{multirow}
\renewcommand{\bottomfraction}{0.45}
\renewcommand{\floatpagefraction}{0.80}
\usepackage{amsmath,amssymb,amsfonts}

\usepackage{bm}
\usepackage{microtype}
\usepackage{xspace}
\usepackage[table]{xcolor}
\usepackage{subcaption}
\usepackage{tikz}
\usetikzlibrary{arrows.meta,positioning,calc,fit,backgrounds,shapes.geometric}
\usepackage{pgfplots}
\pgfplotsset{compat=1.18}
\usepackage{siunitx}
\usepackage{enumitem}
\setlist{nosep,leftmargin=*}
\usepackage{wrapfig}
\usepackage{needspace}
\usepackage{placeins}
\usepackage{float}
\graphicspath{{figures/}}

\definecolor{inkText}{HTML}{2A3342}
\definecolor{inkMuted}{HTML}{6B7484}
\definecolor{accent}{HTML}{8A5CD9}
\definecolor{accentDark}{HTML}{5F3DB0}
\definecolor{encFill}{HTML}{D9F1EA}
\definecolor{encLine}{HTML}{35A08C}
\definecolor{decFill}{HTML}{FDE8E4}
\definecolor{decLine}{HTML}{D97B62}
\definecolor{tokBlue}{HTML}{4E86E8}
\definecolor{tokOrange}{HTML}{E09A3A}
\definecolor{panelFill}{HTML}{F8F9FB}
\definecolor{panelLine}{HTML}{E2E7EE}
\definecolor{good}{HTML}{2E9E6B}
\definecolor{bad}{HTML}{C0392B}
\definecolor{warn}{HTML}{A86B16}

\definecolor{F2F2F2}{HTML}{F2F2F2}
\definecolor{encFillC}{HTML}{9CCDC4}
\definecolor{decFillC}{HTML}{AFA9D3}
\definecolor{critBar}{HTML}{A7C9C9}
\definecolor{latEnd}{HTML}{F1B56D}
\definecolor{latInt}{HTML}{F67B6E}
\definecolor{teachFeat}{HTML}{A6C9EB}
\definecolor{frozenGrey}{HTML}{E8E8E8}
\definecolor{backdrop}{HTML}{F8F9F8}

\newcommand{\ifid}{iFID\xspace}
\newcommand{\rfid}{rFID\xspace}
\newcommand{\gfid}{gFID\xspace}
\newcommand{\lpips}{LPIPS\xspace}
\newcommand{\ours}{InterpTok\xspace}
\newcommand{\maetok}{MAETok\xspace}
\xspaceaddexceptions{\,}

\title{Image Tokenizers Should Be\\Trained Between Images}

\author{Ole Bei{\ss}wenger, Jan-Niklas Dihlmann \& Hendrik Lensch \\
University of T\"ubingen}

\ppfinalcopy

\hypersetup{colorlinks=true, linkcolor=[HTML]{001473}, citecolor=[HTML]{001473}, urlcolor=[HTML]{001473}, pdftitle={Image Tokenizers Should Be Trained Between Images}, pdfauthor={Ole Beißwenger, Jan-Niklas Dihlmann, Hendrik Lensch}}

\makeatletter
\renewcommand\paragraph{\@startsection{paragraph}{4}{\z@}{0.5ex plus 0.3ex minus .2ex}{-1em}{\normalsize\bf}}
\makeatother

\begin{document}
\maketitle

\input{sections/0_abstract}
\input{sections/1_intro}
\input{sections/2_related}
\input{sections/4_method}
\input{sections/5_experiments}
\input{sections/6_conclusion}
\label{end:main}

\subsubsection*{Reproducibility statement}
\input{sections/8_reproducibility}

\subsubsection*{Ethics statement}
\input{sections/9_ethics}

\subsubsection*{AI use statement}
\input{sections/10_ai_use}

\subsubsection*{Acknowledgments}
Funded by the Deutsche Forschungsgemeinschaft (DFG, German Research Foundation) under Germany's Excellence Strategy -- EXC number 2064/1 -- Project number 390727645. This work was supported by the German Research Foundation (DFG): SFB 1233, Robust Vision: Inference Principles and Neural Mechanisms, TP 02, project number: 276693517. This work was supported by the Tübingen AI Center. The authors thank the International Max Planck Research School for Intelligent Systems (IMPRS-IS) for supporting Jan-Niklas Dihlmann.

\bibliography{references}
\bibliographystyle{preprint}

\appendix
\raggedbottom
\makeatletter
\let\itk@subsection\subsection
\renewcommand{\subsection}{\FloatBarrier\itk@subsection}
\let\itk@section\section
\renewcommand{\section}{\FloatBarrier\itk@section}
\newcommand{\subsectionnobarrier}{\itk@subsection}
\newcommand{\sectionnobarrier}{\itk@section}
\makeatother
\setcounter{topnumber}{4}
\setcounter{bottomnumber}{3}
\setcounter{totalnumber}{6}
\renewcommand{\topfraction}{0.95}
\renewcommand{\bottomfraction}{0.70}
\renewcommand{\textfraction}{0.05}
\renewcommand{\floatpagefraction}{0.80}
\makeatletter\setlength{\@fptop}{0pt}\setlength{\@fpsep}{8pt plus 2pt}\setlength{\@fpbot}{0pt plus 1fil}\makeatother
\section*{Overview}
The appendix first shows further results and then the details behind them.
App.~\ref{app:qualitative} shows decoded paths between thirteen image pairs.
App.~\ref{app:results} reports generated samples and generation quality over the generator's training, the latent space of each encoder, reconstruction on ImageNet and MS-COCO, mixtures of several codes, and why generation quality changes with the number of sampling steps.
App.~\ref{app:smallcode} describes an earlier experiment with a much smaller code size, whose decoded mixtures are more semantic compositions of the scenes.
App.~\ref{app:method} gives the implementation details, the training objectives and the evaluation protocol.

\suppressfloats[t]
\input{appendix/C_qualitative}

\input{appendix/F_applications}
\input{appendix/B_additional_results}

\input{appendix/I_mechanism}
\input{appendix/J_controls}
\input{appendix/H_smallcode}
\input{appendix/A_method_details}

\end{document}

%% file: sections/0_abstract.tex
\begin{abstract}
Latent diffusion models generate images as codes of an image tokenizer, whose decoder turns them into the pixel image.
Recent tokenizers improve how well these codes reconstruct an image and how easily a generator learns them.
Yet a generator does not reproduce the codes of training images, and when a generated code lands outside what the decoder expects, the decoded image contains imperfections.
Current methods make the decoder more robust to generated codes by adding noise during training.
However, generated codes lie between the training codes and resemble their mixture more than one of them plus noise.
To improve generated images, decoders should therefore be trained on codes in the space between training samples.
We propose \textbf{\ours}, which fine-tunes \maetok to decode mixtures of image codes into natural images.
Since no target image exists between two training samples, a discriminator judges whether each decoded mixture looks as natural as its two sources.
A mixing ratio loss requires the decoded mixture to keep the mixing ratio of the two codes.
Together, they prevent collapse to a trivial solution and yield a sharp and natural composition of both images.
\ours keeps \maetok's reconstruction quality and improves interpolation quality more than threefold, the best of all tokenizers in a recent benchmark.
Beyond interpolation, \ours improves the generation FID of a matched diffusion transformer from 11.61 to 8.48 without guidance, and even reaches 8.11 at an eighth of the sampling steps, where the generator on \maetok's codes worsens.
The generator trained on our tokenizer also reaches the final result of the one trained on MAETok at half the training steps.
Our findings show that the space between image codes can and should be explicitly supervised rather than ignored.
\end{abstract}

\begin{figure}[!h]
  \centering
  \includegraphics[width=\textwidth]{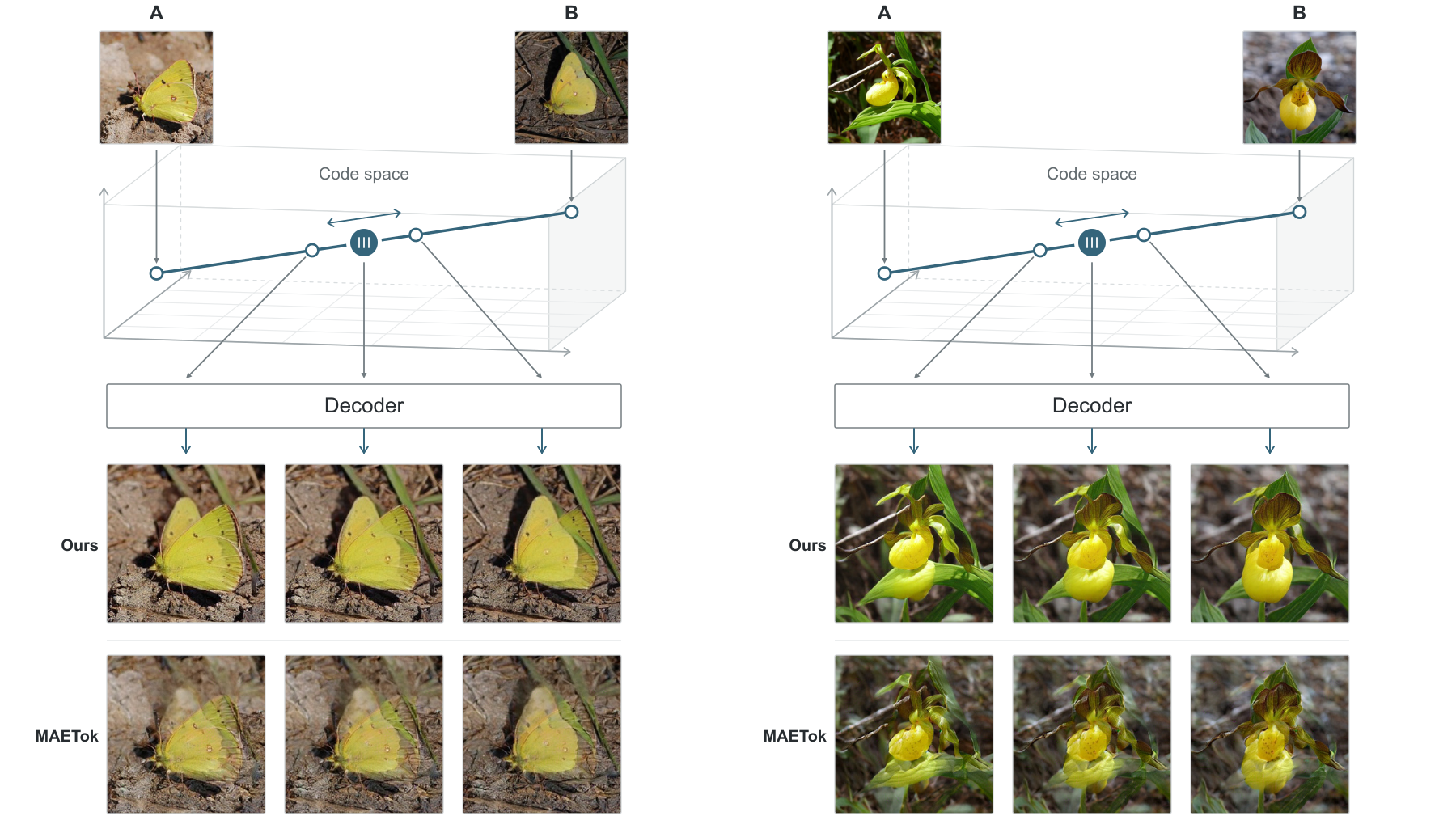}
  \caption{\textbf{Overview.} Reconstruction trains the tokenizer at the codes of images A and B, never on the line between them. \ours trains it there, and the codes on the line between them decode to natural images that change from A toward B, where \maetok superimposes the two images.}
  \label{fig:teaser}
\end{figure}

%% file: sections/1_intro.tex
\section{Introduction}
\label{sec:intro}

Modern image generation largely relies on diffusion and flow matching models~\citep{ho20ddpm,ma24sit}.
These models generate images through many refinement steps, so most operate in a compact latent space to make each step more efficient~\citep{rombach22ldm,peebles23dit}.
In this space, each image is represented by a compact code created by an image tokenizer.
The tokenizer's encoder turns images into codes, and its decoder turns codes back into images.
The generator learns to produce these codes and therefore depends on the space created by the tokenizer.

Improving image generation therefore starts with asking what makes this latent space useful for a generator.
Most tokenizer methods focus on the codes of the training images and make them easier for a generator to learn~\citep{chen25maetok,yao25vavae,fan26uae}.
Standard tokenizer training, however, supervises one code at a time by reconstructing its image.
It tells the decoder what to produce at the codes of the training images, but does not supervise the codes between them.
As Fig.~\ref{fig:teaser} illustrates, a decoder can reconstruct images A and B well and still turn a mixture of their codes into a blurry superposition. 
Reconstruction can therefore give us good endpoints while leaving the path between them undefined.

This gap matters because a generator does not perfectly reproduce training samples but creates codes the decoder never saw.
Recent work shows that generated codes lie near the lines between neighboring training codes.
It also shows that the quality of images decoded on these lines predicts generation quality better than reconstruction quality does~\citep{xu26ifid}.
The decoder must therefore work beyond the exact codes it was trained to reconstruct.
Existing tokenizer and decoder training schemes add noise to training codes and make their local neighborhoods more robust~\citep{yang25ldetok,zheng26rae}.
They improve the visual quality of images decoded around each image code. While this improves generation, it does not target the space between training codes, where generative denoising models tend to place their codes.

In this work, we argue that image tokenizers should be trained between images.
We propose \ours, which mixes the codes of two training images at a known ratio and decodes the result.
Each pair gives us an entire line of training inputs, even though no target image exists along that line.
To prevent trivial solutions such as copying one source, we set two requirements the decoded mixtures must fulfill.
The decoded mixture must look like a natural image, and it must remain tied to its position between the two sources.
Our first component, the \emph{interpolation objective}, addresses both.
An \emph{interpolation discriminator} compares the decoded mixture with one source at a time in the features of a frozen pretrained network and pushes it away from blurry overlays~\citep{caron21dino,sauer21projectedgan}.
A \emph{mixing ratio loss} prevents the decoder from choosing either source and keeps the result at the mixing ratio, in pixels and in the features of a frozen vision model~\citep{simeoni25dinov3}.
Together, they enforce natural-looking images that are true compositions of both sources at the right ratio, allowing supervision without target images.

Interpolation and reconstruction alone do not guarantee a latent space that a generator can learn easily.
In our ablations, the interpolation objective improves the visual fidelity and sharpness of images between codes, but leaving out losses that structure the latent space makes the codes harder for the generator to learn.
We therefore introduce a second component, \emph{regularization losses}.
They align the codes with the features of the same vision model, so that images with similar content receive similar codes~\citep{yao25vavae}. MAETok structures its codes too, through auxiliary decoders that predict pretrained features of masked image parts, and our regularization losses replace them.
\ours combines the interpolation objective and the regularization losses with the original reconstruction objective of \maetok~\citep{chen25maetok}.

Our contributions can be summarized as follows:
\begin{itemize}
\item \textbf{Training Between Images:} We propose a training principle for latent image tokenizers that supervises the space between image codes. An interpolation discriminator and a mixing ratio loss make the decoded mixture of two codes a natural image at the right ratio.
\item \textbf{\ours:} Trained end to end with the interpolation objective and our regularization losses, \ours improves interpolation quality more than threefold and generation quality by more than a quarter and lets its generator train and sample faster, while keeping the reconstruction quality of \maetok.
\item \textbf{Fast Decoder Fine-Tuning:} For \maetok, whose latent space is already structured, fine-tuning only the decoder with our objective is a cheap way to improve the generators already trained on it, without retraining them.
\end{itemize}

%% file: sections/2_related.tex
\section{Related Work}
\label{sec:related}

\textbf{Image tokenizers.}
The objective of image tokenizers is to compress images into a compact latent space and to decode the codes back to images. A generative model can then be trained in this compressed space, improving efficiency and performance~\citep{rombach22ldm,peebles23dit}.
Early tokenizers were plain autoencoders trained to reconstruct their input~\citep{hinton06reducing}, which leaves the shape of their latent space unconstrained.
Variational autoencoders add a penalty that keeps the code distribution close to a Gaussian, which makes the latent space  easier for a generator to learn~\citep{kingma14vae,rombach22ldm}.
Compression loses information, so the optimum of a pure pixel loss is an average of the possible images, which blurs the reconstructions.
VQ-GAN therefore adds a discriminator that judges whether reconstructed images look natural~\citep{esser21taming}.
All of them train the decoder on the codes of single images, and none trains it between two images.
More recent work shapes the latent space itself instead of optimizing reconstruction alone, with 1D codes whose tokens are not tied to patches in the image~\citep{yu24titok} and with codes aligned to the features of pretrained vision models~\citep{yao25vavae}.
\maetok~\citep{chen25maetok} showed that a diffusion model learns the codes more easily when they form fewer and more separated clusters, and reached this structure with a plain autoencoder trained by masked modeling, where auxiliary decoders predict features of the hidden image patches from the codes.
Because masking degrades reconstruction, \maetok then freezes the encoder and fine-tunes only the pixel decoder for a few epochs, which improves fine detail.

\textbf{Behavior of generative denoising models.}
Generative denoising models do not simply recreate the image codes they saw during training. This can lead to generated codes that deviate from the codes the decoder was optimized for. These out-of-distribution codes are then decoded to images with blur or artifacts.
Recent work therefore trains the decoder on codes other than the clean training codes, by transforming the code~\citep{kouzelis25eqvae}, mixing it with noise~\citep{yang25ldetok,zheng26rae} or post-training on the samples of a trained generator~\citep{qiu25posttrain}.
However, \citet{aithal24hallucination} showed that a diffusion model interpolates between nearby modes of its training data and produces samples that exist nowhere in it, which they call mode interpolation.
\citet{xu26ifid} carried this to tokenizer codes, where the optimal denoiser outputs a weighted average of the training codes~\citep{karras22edm}.
They show empirically that the quality of decoded mixtures of image codes (\ifid) predicts the generation quality (\gfid) far better than reconstruction quality (\rfid), which is uncorrelated with it or even negatively correlated.
Generated codes therefore resemble mixtures of image codes.

\textbf{Supervising between images.}
Training an autoencoder on interpolation goes back to representation learning, where a realistic interpolation was taken as a sign that the autoencoder had captured the structure of its data instead of memorizing it~\citep{berthelot18acai}.
ACAI trains the autoencoder so that its decoded mixtures fool a critic that regresses the mixing coefficient~\citep{berthelot18acai}, AMR~\citep{beckham19amr} and \citet{oring21shaping} train it against a discriminator between real and mixed images, and GAIA against an autoencoder that is trained to reconstruct real images but not mixtures~\citep{sainburg18gaia}. ACAI and AMR report smoother interpolations of small images, and codes that are more useful for classification.
Recently, CompTok~\citep{zhao26comptok} argued that a generator learns a token space more easily when the images decoded along the line between the codes of two nearby images stay realistic, and measured this as mode connectivity.
To raise it, CompTok trains on codes formed by swapping token subsets between two images under an adversarial realism term, but a swap composes real token values, whereas a generator produces values between two real ones~\citep{xu26ifid}, and CompTok does not train on the line it measures. The gap between swapped and generated values arises only for continuous tokens, as in \maetok and CompTok, since a generator of discrete tokens only selects real codebook entries.
Complementary to these methods, \ours carries the adversarial interpolation objective to a tokenizer for latent diffusion, trains on the line that CompTok measures, and holds the decoded mixture at the mixing ratio with a mixing ratio loss.

%% file: sections/4_method.tex
\section{Method}
\label{sec:method}

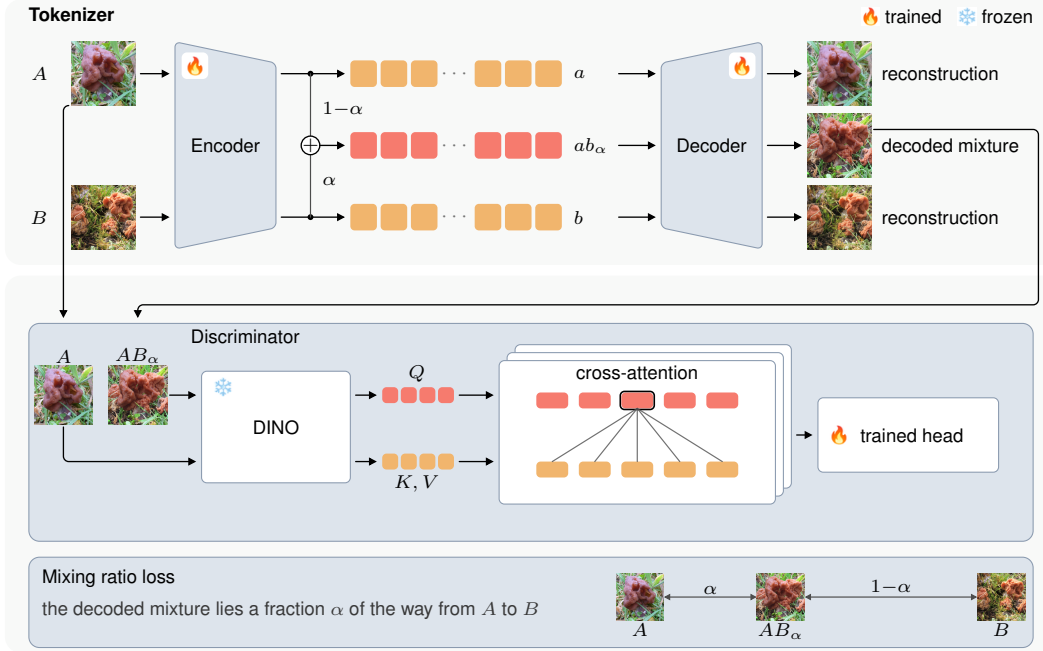
\begin{figure}[t]
  \centering
  \input{figures/fig_method.tex}
  \caption{\textbf{Overview of \ours.} The encoder maps two images $A$ and $B$ to their codes $a$ and $b$, and the decoder maps the codes and their mixture $ab_\alpha$ to two reconstructions and the decoded mixture $AB_\alpha$. The interpolation discriminator judges the decoded mixture beside one of its sources, and the mixing ratio loss keeps it a fraction $\alpha$ of the way from $A$ to $B$.}
  \label{fig:method}
\end{figure}

\ours is an image tokenizer trained to decode mixtures of image codes into sharp, natural images that combine both sources.
It adds two components to the reconstruction objective of \maetok.
The interpolation objective trains the decoder on mixtures of the codes of two images with an interpolation discriminator, which judges each decoded mixture beside one of its two source images, and two mixing ratio losses, which require the decoded mixture to be a composition at the ratio at which the two codes were mixed.
The regularization losses structure the latent space semantically.
Fig.~\ref{fig:method} gives an overview.

\subsection{Background}
\label{sec:method:background}

\textbf{Latent diffusion.}
Latent diffusion models generate images in the compact latent space of an image tokenizer instead of in pixel space, since training generative models on pixels is computationally expensive~\citep{rombach22ldm}.
The tokenizer consists of an encoder $E$ and a decoder $D$.
The encoder maps an image $A$ to its code $a = E(A)$, which is much smaller than the image itself.
The decoder maps a code back to an image.
Both are trained together, so that the decoded image $D(E(A))$ is close to the original image $A$.
A diffusion generator is then trained separately to denoise noisy image codes.
To generate an image, the generator starts from pure noise and iteratively refines it into a code, which $D$ then decodes into the pixel image~\citep{ho20ddpm,ma24sit}.
Diffusion generators tend to generate codes that lie between the codes of training images~\citep{aithal24hallucination,xu26ifid}.
The decoder of the tokenizer, however, is trained only on the codes that the encoder produces for training images, leaving the codes between them unsupervised.

\textbf{\maetok.}
We build on \maetok~\citep{chen25maetok}, which shows that a diffusion model learns codes more easily when a Gaussian mixture model fits them with fewer modes.
\maetok obtains such codes with a plain autoencoder, without the penalty of a variational autoencoder.
Its code consists of 128 vectors, called tokens.
\maetok is trained in two stages.
In the first stage, a large part of every image is masked before it enters the encoder.
Small auxiliary decoders then predict, from the code, the features that pretrained vision models extract from the masked parts, extending masked autoencoders~\citep{he22mae} from pixels to features.
Masking forces the encoder to infer the hidden parts of an image from the visible ones, which leads to such codes.
However, a decoder trained on the codes of partly masked images reconstructs fewer details.
To address this, the second stage freezes the encoder and fine-tunes only the decoder on the codes of complete images, which recovers the lost details.

We choose \maetok for three reasons.
First, we want to show that our method is not limited to the decoder, and therefore choose among tokenizers that train their own encoder.
Among those in the benchmark of \citet{xu26ifid} that do not start from a pretrained vision model, \maetok has the highest interpolation quality.
We attribute this to its masked training, where the decoder is trained to reconstruct the complete image from the code of a partly masked image.
Second, its codes are continuous.
In a quantized tokenizer, every token is an index into a codebook, and there is no code between two indices.
Third, \maetok is a 1D tokenizer~\citep{yu24titok}, whose tokens are not tied to locations in the image, while every token of a 2D tokenizer belongs to one patch.
We want mixtures that interpolate between whole images, not between the patches at each location.

\textbf{Reconstruction objective.}
\maetok trains with the standard tokenizer losses~\citep{esser21taming}, a pixel loss, a perceptual loss~\citep{zhang18lpips} and an adversarial loss.
The pixel loss and the perceptual loss compare the decoded image with the original image, in pixels and in the features of a pretrained network.
The adversarial loss trains the tokenizer against a discriminator that distinguishes real images from reconstructions.
It enforces sharp images, whereas the optimum of a pure pixel loss is a slightly blurred image.
Unlike the other two losses, it needs no target image, since the discriminator judges each image on its own.
We denote these losses together as the reconstruction objective $\mathcal{L}_{\mathrm{rec}}$ and their discriminator as the reconstruction discriminator.\looseness=-1

\subsection{The Space Between Two Codes}
\label{sec:method:space}

We study the codes between the codes of two training images.
Given two training images $A$ and $B$, the sources, we obtain their codes $a = E(A)$ and $b = E(B)$.
We form their mixture at ratio $\alpha \in [0,1]$ and decode it,
\begin{equation}
\label{eq:lerp}
ab_\alpha = (1-\alpha)\,a + \alpha\,b, \qquad AB_\alpha = D(ab_\alpha).
\end{equation}
We call $ab_\alpha$ the mixture and $AB_\alpha$ the decoded mixture.
At $\alpha = 0$ and $\alpha = 1$ the mixture is the code of a source, and in between it moves along the line from $a$ to $b$.
Fig.~\ref{fig:teaser} shows what \maetok's decoder produces on this line.
Toward the middle, the decoded image superimposes both sources and is blurrier than either reconstruction.
\citet{xu26ifid} show that the quality of the images decoded between neighboring training codes predicts generation quality better than reconstruction quality does.
Recent methods add noise to the codes during decoder training~\citep{yang25ldetok,zheng26rae}, which makes the decoder more robust around each training code but does not target the space between codes.

A straightforward alternative is to supervise random codes, so that every code decodes to a realistic image.
However, since real images have few intrinsic dimensions~\citep{pope21intrinsic}, their codes occupy only a small part of the latent space, and random codes mostly lie far outside the training distribution.
Decoding a meaningful image from an arbitrary code is far more ambiguous than decoding one between two image codes.
A mixture has no target image either, but it comes with two sources and a ratio, and \ours uses them in place of a target image.

\textbf{Decodability and learnability.}
Two properties of a tokenizer matter for the quality of the output of a diffusion generator.
We call the visual fidelity and sharpness of the images that the decoder produces from generated codes its decodability.
We call how easily a conditioned generator can learn the codes of training images and tell them apart by their condition, such as their class label, the learnability of the tokenizer.
This is easier when the latent space is structured such that the class of an image can be derived from where its code lies, as \maetok finds.
Decodability is thus mainly a property of the decoder, and learnability of the encoder.
\ours targets decodability with the interpolation objective of Sec.~\ref{sec:method:interp} and learnability with the regularization losses of Sec.~\ref{sec:method:reg}.

\subsection{Interpolation Objective}
\label{sec:method:interp}

The interpolation objective trains the tokenizer on decoded mixtures and targets decodability.
It consists of an interpolation discriminator and two mixing ratio losses.

\textbf{Interpolation discriminator.}
A mixture has no target image, and we therefore supervise the decoded mixture with a discriminator.
Adversarial interpolation was introduced for representation learning with small autoencoders.
ACAI~\citep{berthelot18acai} trains a critic to recover the mixing ratio from the decoded mixture and the autoencoder to hide it, and AMR~\citep{beckham19amr} trains a discriminator that tells real images from decoded mixtures, both on small images such as MNIST, SVHN and CIFAR-10.
Their critics judge the mixture alone, so any realistic image passes, whether or not it relates to the two sources, and neither keeps the mixture at the mixing ratio.
Our interpolation discriminator judges the decoded mixture beside one of its sources and is trained to identify which of the two is the real image. We use it to train a tokenizer for latent diffusion on ImageNet~\citep{deng09imagenet} at $256\times256$ resolution.
For an image $X$ and a second image $Y$, the features of $X$ in a frozen DINO network~\citep{caron21dino} attend to those of $Y$ through trained cross-attention layers, so that every part of $X$ is compared with $Y$. We call this the source conditioning of the discriminator.
A trained head maps the result to the score $s(X \mid Y)$, which is higher for images that look more like real images.
Each decoded mixture is compared with both of its sources in turn.
The discriminator is trained to score the sources above the decoded mixture, and the tokenizer is trained to raise the score of the decoded mixture relative to that of its source, a relativistic objective~\citep{jolicoeur19relativistic},
\begin{equation}
\label{eq:disc}
\mathcal{L}_{\mathrm{adv}} = \frac{1}{2} \sum_{S \in \{A, B\}} \Big[ s\big(S \mid AB_\alpha\big) - s\big(AB_\alpha \mid S\big) \Big].
\end{equation}
The decoder is thus trained to decode mixtures into images that look as real as their sources.

\textbf{Mixing ratio loss.}
\label{sec:method:ratio}
A decoder can fool the interpolation discriminator by trivial solutions such as reproducing one of the two sources.
To overcome this, we introduce a pixel mixing ratio loss, which requires the decoded mixture to be a fraction $\alpha$ of the way from $A$ to $B$.
It compares the distance of the decoded mixture to the reconstruction of each source,
\begin{equation}
\label{eq:ratio}
\mathcal{L}_{\mathrm{ratio}} = \left( \frac{d_A}{d_A + d_B} - \alpha \right)^{2},
\end{equation}
where $d_A = \lVert AB_\alpha - D(a) \rVert_1$ and $d_B = \lVert AB_\alpha - D(b) \rVert_1$ are the pixel distances to $D(a)$ and $D(b)$, the reconstructions at the two ends of the line.
At $\alpha = \tfrac12$, for instance, the decoded mixture has to be as far from one reconstruction as from the other.
Pixel distances capture visual but not semantic differences.
A second version, the feature mixing ratio loss, therefore compares the images in the features $f$ of a second frozen network, DINOv3~\citep{simeoni25dinov3}, which the regularization losses also use.
With $\theta$ the angle between two feature vectors, it requires $\theta\big(f(AB_\alpha), f(A)\big) = \alpha\,\theta\big(f(A), f(B)\big)$ and $\theta\big(f(AB_\alpha), f(B)\big) = (1-\alpha)\,\theta\big(f(A), f(B)\big)$.
A crossfade, which lays the two reconstructions over each other, satisfies the pixel version. The interpolation discriminator penalizes it because it does not look like a natural image.
The mixing ratio loss in turn penalizes trivial solutions such as copies.
Together, they steer the decoder toward sharp images that combine both sources.\looseness=-1

\subsection{Regularization Losses}
\label{sec:method:reg}

The interpolation objective targets decodability but not learnability.
We therefore add regularization losses to structure the latent space semantically.
The frozen DINOv3 model, pretrained on large image collections, describes each image by a feature vector.
Alignment losses make the codes carry the semantic information of these features.
VA-VAE~\citep{yao25vavae} aligns the code of a tokenizer with the patch features of such a model, DINOv2~\citep{oquab24dinov2}, at every image position and requires the similarities between the positions of a code to match those between the features.
REPA~\citep{yu25repa} requires the features to be predictable by a small network from the hidden states of a diffusion generator, and GigaTok~\citep{xiong25gigatok} from the activations inside the decoder of a tokenizer.
\ours uses three alignment losses in this spirit, which require the similarities between the codes of different images to match those between their feature vectors, the feature vector to be predictable from the code, and the features of each image patch to be predictable from the activations inside the decoder.
As a result, images with more similar visual and semantic features receive more similar codes, which these works find easier for a generator to learn.
\maetok ties its codes to pretrained features as well, through auxiliary decoders that predict the features of the masked image parts, and \ours replaces these auxiliary decoders with the three alignment losses.
The representation Fr\'echet loss~\citep{yang26fdloss} was introduced for post-training generators, and to our knowledge it has not been used to train a tokenizer before.
It compares the distribution of the decoded images with that of the training images in the features of three frozen image encoders, SigLIP2~\citep{tschannen25siglip2}, Inception-v3~\citep{szegedy16inception} and MAE~\citep{he22mae}, and minimizes the Fr\'echet distance between the two.
The alignment losses act on single images and the representation Fr\'echet loss on a subsample of the decoded images, mixtures included, and together they form $\mathcal{L}_{\mathrm{reg}}$.\looseness=-1

\subsection{Full Objective}
\label{sec:method:objective}

The full objective of \ours combines the reconstruction objective on single images, the interpolation objective on mixtures and the regularization losses,
\begin{equation}
\label{eq:objective}
\mathcal{L} = \mathcal{L}_{\mathrm{rec}} + \mathcal{L}_{\mathrm{adv}} + \mathcal{L}_{\mathrm{ratio}} + \mathcal{L}_{\mathrm{reg}},
\end{equation}
where $\mathcal{L}_{\mathrm{ratio}}$ includes both mixing ratio losses and App.~\ref{app:method:losses} gives the weight of every term.
To give the decoder capacity for the codes between images, we append six blocks to it, as App.~\ref{app:method:growth} describes.
Sec.~\ref{sec:ablations} shows that the gains of the decoder do not depend on this added capacity.

%% file: figures/fig_method.tex
\begingroup%
\providecolor{backdrop}{HTML}{F8F9F8}%
\providecolor{latEnd}{HTML}{F1B56D}%
\providecolor{latInt}{HTML}{F67B6E}%
\providecolor{frozenGrey}{HTML}{E8E8E8}%
\providecolor{blkFill}{HTML}{DDE4EC}%
\providecolor{blkLine}{HTML}{8A9AAF}%
\def\FMthumb#1#2#3#4#5{%
  \IfFileExists{figures/#1}%
    {\node[anchor=south west,inner sep=0pt] at (#3,#4){\includegraphics[width=#5pt,height=#5pt]{#1}};}%
    {\IfFileExists{figures/#2}%
       {\node[anchor=south west,inner sep=0pt] at (#3,#4){\includegraphics[width=#5pt,height=#5pt]{#2}};}%
       {\fill[frozenGrey] (#3,#4) rectangle ++(#5,#5);}}%
}%
\def\FMglyph#1#2#3{%
  \IfFileExists{figures/#1}{\node[anchor=center,inner sep=0pt] at (#2){\includegraphics[width=#3pt]{#1}};}{}%
}%
\def\FMmark#1#2#3#4{%
  \fill[white,rounded corners=2pt] (#2-5,#3-5) rectangle (#2+5,#3+5);%
  \FMglyph{#1}{#2,#3}{#4}%
}%
\begin{tikzpicture}[x=1pt,y=1pt,line cap=round,line join=round,
  every node/.style={inner sep=0pt,outer sep=0pt},
  ptitle/.style={font=\sffamily\bfseries\fontsize{7}{8.4}\selectfont,text=black},
  lab/.style={font=\sffamily\fontsize{7}{8.4}\selectfont,text=black},
  labq/.style={font=\sffamily\fontsize{7}{8.4}\selectfont,text=black!78},
  mlab/.style={font=\fontsize{7}{8.4}\selectfont,text=black},
  dots/.style={font=\fontsize{7}{8.4}\selectfont,text=black!55},
  tile/.style={rounded corners=1.6pt,draw=none},
  ar/.style={draw=black,line width=0.5pt,-{Triangle[width=3pt,length=3pt]}},
  wire/.style={draw=black!70,line width=0.45pt}
  ]
\useasboundingbox (0,-94) rectangle (396,152);

\fill[backdrop,rounded corners=6pt] (0,52) rectangle (396,152);
\fill[backdrop,rounded corners=6pt] (0,-94) rectangle (396,48);

\node[ptitle,anchor=north west] at (9,149) {Tokenizer};
\node[lab,anchor=east] (lgF) at (387,145.5) {frozen};
\FMglyph{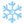}{$(lgF.west)+(-5.5,0)$}{7}
\node[lab,anchor=east] (lgT) at ($(lgF.west)+(-15,0)$) {trained};
\FMglyph{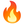}{$(lgT.west)+(-5.5,0)$}{7}

\FMthumb{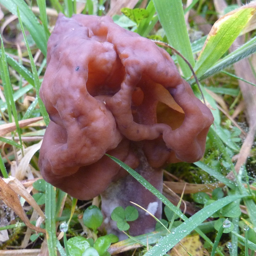}{thumb_A.png}{25}{112}{24}
\FMthumb{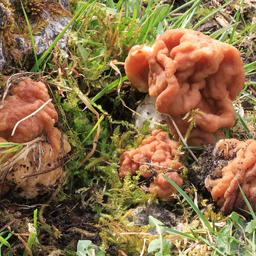}{thumb_B.png}{25}{58}{24}
\node[mlab,anchor=east] at (16,124) {$A$};
\node[mlab,anchor=east] at (16,70) {$B$};
\draw[ar] (51,124) -- (62,124);
\draw[ar] (51,70) -- (62,70);

\filldraw[fill=blkFill,draw=blkLine,line width=0.4pt,rounded corners=2pt] (64,58) -- (102,66) -- (102,128) -- (64,136) -- cycle;
\FMmark{glyph_fire_flat.png}{71.5}{127.0}{7}
\node[lab] at (83.0,97) {Encoder};

\draw[ar] (104,124) -- (128,124);
\draw[ar] (104,70) -- (128,70);
\fill[black] (115,124) circle (1.1);
\fill[black] (115,70) circle (1.1);
\draw[wire] (115,124) -- (115,100.5);
\draw[wire] (115,70) -- (115,93.5);
\filldraw[fill=white,draw=black,line width=0.5pt] (115,97) circle (3.5);
\node[mlab] at (115,97.2) {$+$};
\draw[ar] (118.5,97) -- (128,97);
\node[mlab,anchor=west] at (119.5,110.5) {$1{-}\alpha$};
\node[mlab,anchor=west] at (119.5,83.5) {$\alpha$};

\foreach \x in {130.0,141.4,152.8,176.8,188.2,199.6}{%
  \fill[latEnd,tile] (\x,119.0) rectangle ++(10.0,10.0);%
  \fill[latInt,tile] (\x,92.0) rectangle ++(10.0,10.0);%
  \fill[latEnd,tile] (\x,65.0) rectangle ++(10.0,10.0);}
\node[dots] at (169.8,124) {$\cdots$};
\node[dots] at (169.8,97) {$\cdots$};
\node[dots] at (169.8,70) {$\cdots$};
\node[mlab,anchor=west] at (214.1,124) {$a$};
\node[mlab,anchor=west] at (214.1,97) {$ab_\alpha$};
\node[mlab,anchor=west] at (214.1,70) {$b$};

\draw[ar] (231,124) -- (245,124);
\draw[ar] (231,97) -- (245,97);
\draw[ar] (231,70) -- (245,70);
\filldraw[fill=blkFill,draw=blkLine,line width=0.4pt,rounded corners=2pt] (247,66) -- (285,58) -- (285,136) -- (247,128) -- cycle;
\FMmark{glyph_fire_flat.png}{277.5}{127.0}{7}
\node[lab] at (266.0,97) {Decoder};
\draw[ar] (287,124) -- (300,124);
\draw[ar] (287,97) -- (300,97);
\draw[ar] (287,70) -- (300,70);

\FMthumb{thumb_recA.png}{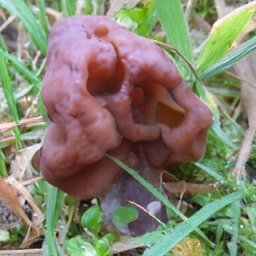}{302}{112}{24}
\FMthumb{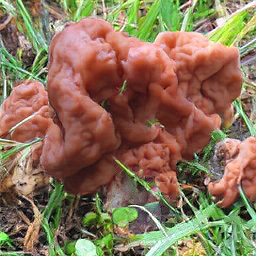}{thumb_int.png}{302}{85}{24}
\FMthumb{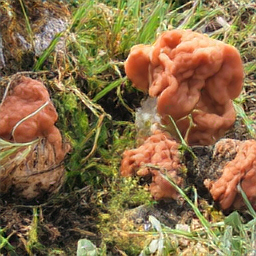}{thumb_rec_B.png}{302}{58}{24}
\node[lab,anchor=west] at (330,124) {reconstruction};
\node[lab,anchor=west] at (330,97) {decoded mixture};
\node[lab,anchor=west] at (330,70) {reconstruction};

\draw[ar,rounded corners=2pt] (24,112) -- (22,112) -- (22,32);
\draw[ar,rounded corners=2pt] (327,103) -- (389,103) -- (389,37) -- (50,37) -- (50,32);

\filldraw[fill=blkFill,draw=blkLine,line width=0.4pt,rounded corners=4pt] (9,-52) rectangle (387,30);
\node[lab,anchor=west] at (70,25) {Discriminator};

\FMthumb{thumb_A.png}{thumb_A.png}{11}{-8}{22}
\FMthumb{thumb_int.png}{thumb_int.png}{39}{-8}{22}
\node[mlab,anchor=south] at (22,15) {$A$};
\node[mlab,anchor=south] at (50,15) {$AB_\alpha$};

\draw[ar] (62,3) -- (72,3);
\draw[ar,rounded corners=2pt] (22,-9) -- (22,-22) -- (72,-22);
\filldraw[fill=white,draw=blkLine,line width=0.4pt,rounded corners=2pt] (74,-30) rectangle (130,12);
\FMmark{glyph_snow_flat.png}{82.0}{6.0}{7}
\node[lab] at (102,-9) {DINO};

\draw[ar] (132,3) -- (140,3);
\draw[ar] (132,-22) -- (140,-22);
\foreach \x in {142,149,156,163}{\fill[latInt,tile] (\x,0) rectangle ++(6,6);}
\foreach \x in {142,149,156,163}{\fill[latEnd,tile] (\x,-25) rectangle ++(6,6);}
\node[mlab,anchor=south] at (155,8) {$Q$};
\node[mlab,anchor=north] at (155,-27) {$K,V$};
\draw[ar] (171,3) -- (184,3);
\draw[ar] (171,-22) -- (184,-22);

\filldraw[fill=white,draw=blkLine,line width=0.4pt,rounded corners=2pt] (192,-32) rectangle (296,22);
\filldraw[fill=white,draw=blkLine,line width=0.4pt,rounded corners=2pt] (189,-35) rectangle (293,19);
\filldraw[fill=white,draw=blkLine,line width=0.4pt,rounded corners=2pt] (186,-38) rectangle (290,16);
\node[lab,anchor=north] at (238,14) {cross-attention};
\foreach \x in {200,216,232,248,264}{\fill[latInt,tile] (\x,-2) rectangle ++(12,6);}
\foreach \x in {200,216,232,248,264}{\fill[latEnd,tile] (\x,-28) rectangle ++(12,6);}
\draw[draw=black,line width=0.6pt,rounded corners=1.6pt] (231.6,-2.4) rectangle (244.4,4.4);
\foreach \x in {206,222,238,254,270}{\draw[wire,opacity=0.8] (238,-2) -- (\x,-22);}

\draw[ar] (298,-12) -- (304,-12);
\filldraw[fill=white,draw=blkLine,line width=0.4pt,rounded corners=2pt] (306,-26) rectangle (374,2);
\FMmark{glyph_fire_flat.png}{314.0}{-12.0}{7}
\node[lab,anchor=west] at (322,-12) {trained head};

\filldraw[fill=blkFill,draw=blkLine,line width=0.4pt,rounded corners=4pt] (9,-92) rectangle (387,-58);
\node[lab,anchor=west] at (14,-66) {Mixing ratio loss};
\node[labq,anchor=west] at (14,-78) {the decoded mixture lies a fraction $\alpha$ of the way from $A$ to $B$};
\draw[wire,{Triangle[width=2.4pt,length=2.4pt]}-{Triangle[width=2.4pt,length=2.4pt]}] (248,-73) -- (283,-73);
\draw[wire,{Triangle[width=2.4pt,length=2.4pt]}-{Triangle[width=2.4pt,length=2.4pt]}] (301,-73) -- (366,-73);
\FMthumb{thumb_A.png}{thumb_A.png}{230}{-82}{18}
\FMthumb{thumb_int.png}{thumb_int.png}{283}{-82}{18}
\FMthumb{thumb_B.png}{thumb_B.png}{366}{-82}{18}
\node[mlab,anchor=north] at (239,-83) {$A$};
\node[mlab,anchor=north] at (292,-83) {$AB_\alpha$};
\node[mlab,anchor=north] at (375,-83) {$B$};
\node[mlab,anchor=south] at (265.5,-72) {$\alpha$};
\node[mlab,anchor=south] at (333.5,-72) {$1{-}\alpha$};
\end{tikzpicture}%
\endgroup%

%% file: sections/5_experiments.tex
\section{Experiments}
\label{sec:experiments}

We evaluate \ours on reconstruction, interpolation, generation and learnability, compare it with \maetok and with the tokenizers benchmarked by \citet{xu26ifid}, and ablate its components.
Our experiments validate three claims.
First, a modern tokenizer can be trained between images together with current regularization losses. \ours improves interpolation quality, generation quality and learnability while keeping reconstruction quality.
Second, the regularization losses alone do not improve generation, and trained with them but without interpolation, the tokenizer has the worst gFID of all variants.
Third, a conditioned generator needs both learnability, a property of the encoder, and decodability, a property of the decoder. Most of the \ours generation gain compared to \maetok comes from the decoder.

\subsection{Setup}
\label{sec:setup}

\textbf{Implementation details.}
We start from the released weights of \maetok and train its encoder and decoder together on ImageNet~\citep{deng09imagenet} at $256\times256$ resolution.
The interpolation objective replaces \maetok's masking as a self-supervised objective.
We evaluate on ImageNet and MS-COCO~\citep{lin14coco}.
For generation, we train a SiT-B~\citep{ma24sit} conditioned on class labels with the same setting on the codes of each tokenizer.
We also decode the generated codes of the SiT-XL released with \maetok~\citep{chen25maetok} with our decoder trained on \maetok's frozen encoder.
We sample every generator without classifier-free guidance~\citep{ho22cfg}, whose strength would have to be tuned for each tokenizer.
More details are provided in App.~\ref{app:method}.

\input{tables/T1_main.tex}
\textbf{Metrics.}
For reconstruction, we compare each decoded image with its original, using the Peak Signal-to-Noise Ratio (PSNR) for pixel accuracy, the Structural Similarity Index (SSIM)~\citep{wang04ssim} for structural preservation and the Learned Perceptual Image Patch Similarity (LPIPS)~\citep{zhang18lpips} for perceptual alignment.
Decoded mixtures have no original to compare with.
The Fr\'echet Inception Distance (FID)~\citep{heusel17fid}, the standard metric for the quality of generated images, needs none, since it compares the distribution of a model's images with that of images of the dataset.
Following \citet{xu26ifid}, we therefore measure decoded mixtures with the interpolated FID (iFID).
iFID decodes the midpoint of a spherical path between each validation code and one of its nearest training codes and computes the FID of these images.
It thus serves as our measure of decodability.
We also report the FID of tokenizer reconstructions (rFID) and of generated images (gFID).
We measure gFID at the usual 250 sampling steps and at fewer, since fewer steps make sampling cheaper.
For learnability, we conduct linear probing as \maetok does, training a linear classifier to predict the class of an image from its code, and report its accuracy.
ImageNet labels every image with a class, and the generator is conditioned on this class label. We also report how often a code lies closest to the average code of its own class rather than to the average code of a different class.

\subsection{Main Results}
\label{sec:main_results}

\textbf{Interpolation.}
\ours sets the best interpolation quality in the comparison.
In Table~\ref{tab:main}, it reduces the iFID of \maetok more than threefold, from 14.42 to 4.30, lower than that of every tokenizer in the comparison released by \citet{xu26ifid}.
On random MS-COCO pairs, the FID of its decoded midpoints is six times lower than \maetok's, as App.~\ref{app:results:ood} shows.
This demonstrates that decoding between codes can be learned, and App.~\ref{app:qualitative} shows decoded paths for thirteen image pairs.
The next best, RAE~\citep{zheng26rae}, whose code consists of the features of the pretrained network DINOv2, reaches 1.8 times our iFID.
RAE and l-DeTok~\citep{yang25ldetok} train their decoders on noisy codes, indicating that robustness to noise improves interpolation quality less than training between codes, as the decoder with noise augmentation of Table~\ref{tab:abl} confirms.

\begin{wrapfigure}[18]{r}{0.36\textwidth}
\vspace{-\intextsep}
\centering
\includegraphics[width=0.33\textwidth]{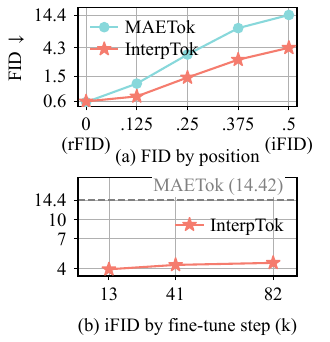}
\caption{\textbf{FID between two codes.} (a) From a code to its midpoint with one of its nearest training codes, on a logarithmic axis. (b) The iFID of \ours over its fine-tuning steps.}
\label{fig:analysis}
\end{wrapfigure}

\textbf{Between the codes.}
Fig.~\ref{fig:analysis}a shows the FID of the images that \maetok and \ours decode from a code to its midpoint with one of its nearest training codes.
At the code, both reach about the same FID.
At the midpoint, \maetok's FID is 25 times its FID at the code, and \ours's only 7 times.
Their decoders thus differ most between the codes, where reconstruction metrics are not computed.
Fig.~\ref{fig:analysis}b shows that \ours reaches most of its gain at the midpoint within the first 13{,}000 fine-tuning steps.

\Needspace{19\baselineskip}
\textbf{Generation.}
\ours makes generation better, cheaper to sample and faster to train.
In Table~\ref{tab:nfe}, replacing \maetok with \ours reduces the gFID of the generator from 11.61 to 8.48, in line with \citet{xu26ifid}, who find that iFID predicts gFID.
With an eighth of the sampling steps, and thus an eighth of the sampling cost, \ours's generator reaches an even lower gFID, while the gFID of the generator on \maetok's codes rises with every reduction of the steps.
\ours's generator also reaches the final gFID of \maetok's generator after only half of the training steps. App.~\ref{app:results:gfid} shows the gFID over the generator's training, and Fig.~\ref{fig:gen_cfg1} shows generated samples.

\textbf{Reconstruction.}
Training between images does not cost reconstruction quality.
In Table~\ref{tab:recon}, the rFID of \ours on ImageNet stays within three percent of \maetok's, and its PSNR, SSIM and LPIPS improve on both ImageNet and MS-COCO, while its rFID on MS-COCO is higher.

\textbf{Learnability.}
\ours's codes are more discriminative than \maetok's, although \maetok also ties its codes to pretrained features.
In Table~\ref{tab:abl}, a linear classifier predicts the class from \ours's codes more accurately, and App.~\ref{app:mechanism} shows that 73 percent of \ours's codes lie closest to the average code of their own class, against 55 percent of \maetok's.
The class of an image can thus be derived from its code more reliably, and a model conditioned on class labels can learn more easily which codes belong to which class. App.~\ref{app:results:probe} shows the codes of both encoders.\looseness=-1

\subsection{Ablations}
\label{sec:ablations}

\input{tables/T6_nfe.tex}
Table~\ref{tab:abl} compares variants of \maetok and of \ours, all trained on the same data.
In the upper rows, two variants add interpolation or the regularization losses to \maetok, and one variant of \ours removes the three alignment losses and the feature mixing ratio loss and keeps the representation Fr\'echet loss.
In the lower rows, only the decoder is trained, with \maetok's encoder frozen, and each row marked w/o removes one part of our objective.

\input{tables/T3_design.tex}
\textbf{Interpolation and regularization.}
Trained with interpolation but without the regularization losses, the tokenizer improves interpolation quality nearly as much as \ours, but a linear classifier predicts the class from its codes less accurately than from \maetok's.
Trained with the regularization losses but without interpolation, it raises the probe accuracy but barely improves interpolation quality, and its gFID is the highest of all variants.
With only interpolation or only the regularization losses, the tokenizer has a higher gFID than \maetok, while \ours, which combines both, reaches a lower one, indicating that a generator needs both learnability and decodability.
Without the alignment losses and the feature mixing ratio loss, \ours keeps most of its interpolation gain, but its probe accuracy falls below \maetok's and its gFID rises far above \maetok's, indicating that most of the learnability comes from these losses.

\textbf{Decoder alone.}
The decoder-only variant of \ours keeps \maetok's codes by freezing the encoder, so \maetok's generators can use it without retraining.
In Table~\ref{tab:nfe}, the SiT-B on \maetok's codes nearly reaches the gFID of \ours's own generator with our decoder, and the released SiT-XL improves at every number of sampling steps.
Most of the generation gain thus comes from the decoder, and it comes mostly from training between codes rather than from training longer.
In Table~\ref{tab:abl}, for the SiT-B on \maetok's codes and after 13{,}000 steps each, our decoder lowers the gFID from 11.61 to 8.43.
A decoder trained just as long with only \maetok's own losses lowers it to 10.80.
When we add only our interpolation discriminator to \maetok's losses, the gFID already drops to 9.71.
Adding noise to codes when training the decoder, as in \citet{zheng26rae}, lowers it to 8.83, but leaves interpolation quality almost unchanged and more than doubles the rFID.

\textbf{Components of the objective.}
To find which parts of the objective cause the gain, each row marked w/o in the lower block of Table~\ref{tab:abl} removes exactly one part from our decoder trained for 13{,}000 steps and keeps all the others.
The six appended blocks add capacity, and removing them changes neither iFID nor gFID by more than 0.02, so the gain does not come from added capacity.
The representation Fr\'echet loss matches the feature statistics of decoded and real images, and removing it raises the SiT-B gFID only from 8.43 to 8.56, so most of the gain comes from the other losses.
Table~\ref{tab:abl} and App.~\ref{app:results:swap_xl} compare our 13{,}000-step decoder with noise augmentation as in \citet{zheng26rae}, and ours is better in interpolation, reconstruction and generation with both generators.
Removing the feature mixing ratio loss costs the most and raises iFID from 4.48 to 5.23 and gFID from 8.43 to 8.90, removing the source conditioning raises only iFID to 4.69, and removing the pixel mixing ratio loss or the Inception part of the Fr\'echet loss changes neither by more than 0.13.

\textbf{Latent space.}
In Table~\ref{tab:mech_latent}, with the regularization losses but without interpolation, a code's nearest neighbor belongs to its own class twice as often as with \maetok's encoder, but the codes spread over three times as many effective dimensions.
The effective dimensions are the participation ratio of the covariance spectrum, which counts the directions that carry most of the variance.
With \ours's encoder, the nearest neighbor belongs to its own class about as often as with this variant, while the codes spread over a third fewer effective dimensions than this variant's.
A generator models every direction along which the codes vary, and at a comparable class structure, a latent space with fewer effective dimensions is easier for a diffusion model to learn~\citep{zhong26diffusability,yao25vavae}.

\textbf{Fewer sampling steps.}
With the sampler of Table~\ref{tab:nfe}, \ours's generator reaches a lower gFID with 64 or 32 sampling steps than with 250, while every other generator we sample with fewer steps worsens, including \maetok's generator with our decoder and the generators of the variants trained without interpolation and without the regularization losses, as App.~\ref{app:mechanism} shows.
We attribute this to the combination of our latent space with a decoder that decodes the space between codes into sharp images.
With fewer and larger steps, a generated code ends closer to the average code of its class, and the generated codes of a class end closer together.
With \ours's encoder, the codes near a class average belong to that class more than twice as often as with \maetok's, so such a code still shows its class.
\ours's decoder, trained between codes, also decodes averages of codes of one class with a lower FID than \maetok's decoder.
Of all these generators, only \ours's combines both properties, and only its gFID falls with fewer steps.

%% file: tables/T1_main.tex
\Needspace{24\baselineskip}
\begin{wraptable}[25]{r}{0.55\textwidth}
\vspace{-\intextsep}
\caption{\textbf{Reconstruction and interpolation on ImageNet.} \ours against nine of the thirteen tokenizers compared by \citet{xu26ifid}, and Table~\ref{tab:sources} lists the omitted rows. Among the nine are SD-VAE~\citep{rombach22ldm}, EQ-VAE~\citep{kouzelis25eqvae}, DM-VAE~\citep{ye26dmvae}, REPA-E~\citep{leng25repae} and SoftVQ-VAE~\citep{chen25softvq}. Latent dims counts the values of a code. Bold is the best value among trained encoders, and underlining marks an improvement of \ours over \maetok. $^{\dagger}$Quoted from that comparison or the tokenizer's own paper.}
\label{tab:main}
\centering\footnotesize\setlength{\tabcolsep}{4pt}
\begin{tabular}{@{}p{2.55cm}crr@{}}
\toprule
Tokenizer & Latent dims & \rfid\,$\downarrow$ & \ifid\,$\downarrow$ \\
\midrule
\multicolumn{4}{@{}l}{\textit{DINOv2 encoder}} \\
DM-VAE & 8{,}192 & 0.683 & 8.29 \\
RAE & 196{,}608 & 0.57$^{\dagger}$ & 7.68$^{\dagger}$ \\
\midrule
\multicolumn{4}{@{}l}{\textit{Trained encoder}} \\
SD-VAE & 4{,}096 & 0.62$^{\dagger}$ & 59.91$^{\dagger}$ \\
EQ-VAE & 4{,}096 & 0.82$^{\dagger}$ & 47.51$^{\dagger}$ \\
REPA-E & 4{,}096 & -- & 36.70$^{\dagger}$ \\
SoftVQ-VAE & 2{,}048 & 0.61$^{\dagger}$ & 26.91$^{\dagger}$ \\
VA-VAE & 8{,}192 & \textbf{0.28$^{\dagger}$} & 19.57$^{\dagger}$ \\
l-DeTok & 4{,}096 & 0.68$^{\dagger}$ & 17.51$^{\dagger}$ \\
\maetok & 4{,}096 & 0.582 & 14.42 \\
\ours & 4{,}096 & 0.597 & \underline{\textbf{4.30}} \\
\bottomrule
\end{tabular}
\end{wraptable}

%% file: tables/T6_nfe.tex
\begin{table}[t]
\centering
\begin{minipage}[c]{0.44\textwidth}
\caption{\textbf{Generation against sampling steps.} Each SiT-B is class-conditional, trained for 400{,}000 steps and sampled without guidance with the stochastic sampler of SiT~\citep{ma24sit}. The second row decodes the first row's generated codes, from the same noise, with the 41{,}000-step decoder that Table~\ref{tab:abl} trains on \maetok's frozen encoder. The SiT-XL rows decode samples of the generator released by \citet{chen25maetok}. $n = 50{,}000$ per cell. Bold is the best value in its column and block.}
\label{tab:nfe}
\end{minipage}\hfill
\begin{minipage}[c]{0.52\textwidth}
\centering\footnotesize\setlength{\tabcolsep}{4pt}
\begin{tabular}{@{}lrrrr@{}}
\toprule
\gfid\,$\downarrow$ & 250 & 64 & 32 & 16 \\
\midrule
\multicolumn{5}{@{}l}{\emph{SiT-B}} \\
\maetok & 11.61 & 12.33 & 13.88 & 20.56 \\
\quad w/ our decoder & 8.56 & 8.99 & 9.92 & 15.33 \\
\ours & \textbf{8.48} & \textbf{8.09} & \textbf{8.11} & \textbf{12.48} \\
\midrule
\multicolumn{5}{@{}l}{\emph{SiT-XL released with \maetok}} \\
\maetok & 2.46 & 2.78 & 3.46 & 7.52 \\
\quad w/ our decoder & \textbf{2.26} & \textbf{2.45} & \textbf{2.79} & \textbf{5.80} \\
\bottomrule
\end{tabular}
\end{minipage}
\end{table}

%% file: tables/T3_design.tex
\begin{table}[t]
\caption{\textbf{Ablations of \ours.} The upper rows train encoder and decoder. The lower rows train only the decoder on \maetok's frozen encoder and are evaluated with \maetok's generator. Interpolation is the interpolation discriminator with the pixel mixing ratio loss, reg.\ the regularization losses, DINOv3 the alignment losses and the feature mixing ratio loss, and Inception the Inception-v3 features of the representation Fr\'echet loss. Bold is the best value in its column and block.}
\label{tab:abl}\label{tab:ladder}\label{tab:components}
\centering\small
\begin{tabular*}{\linewidth}{@{\extracolsep{\fill}}lrrrrrr@{}}
\toprule
 & Params (M) & Steps & \ifid\,$\downarrow$ & \rfid\,$\downarrow$ & probe\,$\uparrow$ & \gfid\,$\downarrow$ \\
\midrule
\maetok & 174 & -- & 14.42 & 0.582 & 73.0 & 11.61 \\
\midrule
\multicolumn{7}{@{}l}{\emph{encoder and decoder trained}} \\
\quad w/ interpolation & 174 & 41k & 5.44 & 0.556 & 63.5 & 18.90 \\
\quad w/ reg. & 217 & 41k & 12.86 & 0.588 & 77.0 & 21.15 \\
\quad \ours & 217 & 41k & \textbf{4.30} & 0.597 & \textbf{77.9} & \textbf{8.48} \\
\qquad w/o DINOv3 & 217 & 41k & 4.53 & \textbf{0.544} & 65.8 & 17.58 \\
\midrule
\multicolumn{7}{@{}l}{\emph{decoder only}} \\
\quad \maetok's losses & 174 & 13k & 12.48 & \textbf{0.572} & 73.0 & 10.80 \\
\quad w/ interpolation discriminator & 174 & 13k & 9.06 & 0.634 & 73.0 & 9.71 \\
\quad w/ noise augmentation & 174 & 13k & 12.62 & 1.460 & 73.0 & 8.83 \\
\quad \ours & 217 & 41k & 4.46 & 0.637 & 73.0 & 8.56 \\
\quad \ours & 217 & 13k & 4.48 & 0.616 & 73.0 & 8.43 \\
\qquad w/o six appended blocks & 174 & 13k & 4.47 & 0.614 & 73.0 & \textbf{8.41} \\
\qquad w/o repr.\ Fr\'echet loss & 217 & 13k & \textbf{4.33} & 0.599 & 73.0 & 8.56 \\
\qquad w/o Inception & 217 & 13k & 4.35 & 0.602 & 73.0 & 8.54 \\
\qquad w/o pixel mixing ratio loss & 217 & 13k & 4.49 & 0.615 & 73.0 & 8.45 \\
\qquad w/o feature mixing ratio loss & 217 & 13k & 5.23 & 0.632 & 73.0 & 8.90 \\
\qquad w/o source conditioning & 217 & 13k & 4.69 & 0.607 & 73.0 & 8.44 \\
\bottomrule
\end{tabular*}
\end{table}

%% file: sections/6_conclusion.tex
\section{Conclusion}
\label{sec:conclusion}

Diffusion generators tend to generate codes between the codes of training images, and the quality of the images decoded there predicts generation quality~\citep{xu26ifid}.
We presented \ours, which trains the tokenizer there, on mixtures of the codes of two training images, with an interpolation discriminator, a mixing ratio loss and regularization losses that structure the latent space semantically.

Starting from \maetok, \ours improves interpolation quality more than threefold, beyond every tokenizer in the comparison released by \citet{xu26ifid}, and lowers the gFID by more than a quarter while keeping reconstruction quality.
Its generator reaches an even lower gFID with an eighth of the sampling steps and the final gFID of \maetok's generator after half of the training steps, and its decoder alone improves \maetok's existing generators without retraining them.
Our ablations show that a conditioned generator needs both learnability, a property of the encoder, and decodability, a property of the decoder.

Training between images opens several directions for future work.
App.~\ref{app:smallcode} reports an earlier experiment in which a much smaller code trained with the interpolation objective gave more semantic interpolations.
Decoders are much faster than generative models, so future work could investigate this self-supervised training objective for real-time animation, video generation or granular editing.

Our findings show that the space between image codes can and should be explicitly supervised rather than ignored.

%% file: sections/8_reproducibility.tex
\label{app:ai}%
App.~\ref{app:method} and Table~\ref{tab:config} give every hyperparameter, loss and schedule. \ifid and ImageNet \rfid are measured with the evaluation implementation released by \citet{xu26ifid} at its defaults, Table~\ref{tab:sources} names the other evaluators, and Table~\ref{tab:protocol} lists the main datasets and pairing rules. The implementation of training and evaluation and the configuration of every run are available at \url{https://github.com/cgtuebingen/InterpTok}, and the tokenizer weights of the 41{,}000-step and 82{,}000-step checkpoints of App.~\ref{app:results:telemetry} at \url{https://huggingface.co/CGTuebingen/InterpTok-Models}.

%% file: sections/9_ethics.tex
This work improves image tokenizers for generation and interpolation. We train on ImageNet, and the earlier tokenizer of App.~\ref{app:smallcode} also on ImgEdit~\citep{ye25imgedit}, and we evaluate on ImageNet and MS-COCO
under their research terms and collect no human-subject data. The images at the ends of the paths in
App.~\ref{app:smallcode} and at the corners of its squares were generated by a text-to-image model.

%% file: sections/10_ai_use.tex
Large language model assistants supported this work in finding related work, in the research, in writing code and in drafting and editing
the text. We reviewed everything they produced and take responsibility for the content.

%% file: appendix/C_qualitative.tex
\FloatBarrier
\section{Additional Visualizations}
\label{app:qualitative}
\label{app:qual}
\label{app:qualitative:rule}
\label{app:qualitative:veto}
\label{app:qualitative:corpus}

\begin{figure}[tbp]
  \centering
    \includegraphics[width=0.82\linewidth]{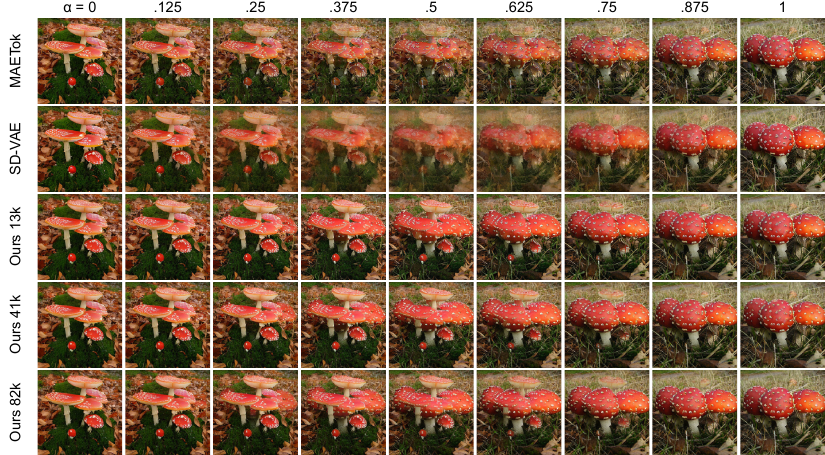}%
  
  \caption{\textbf{Selected interpolation, agaric.}}
  \label{fig:c_pair_01}
\end{figure}

\begin{figure}[!tbp]
  \centering
    \includegraphics[width=0.82\linewidth]{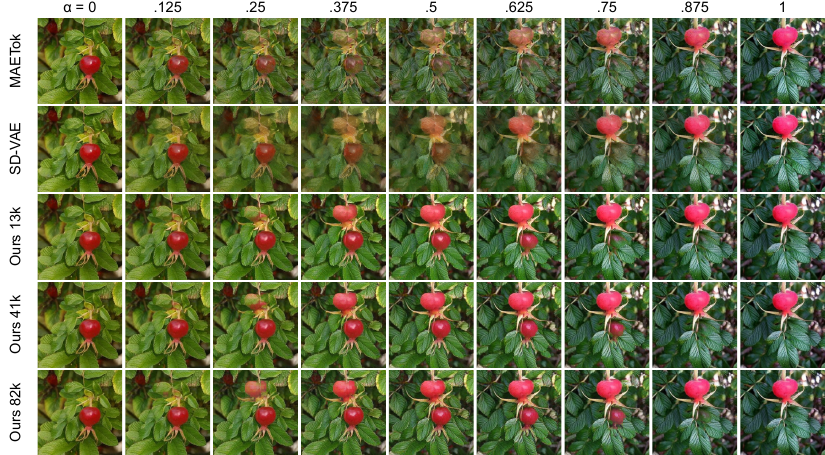}%
  
  \caption{\textbf{Selected interpolation, rose hip.}}
  \label{fig:c_pair_02}
\end{figure}

\begin{figure}[!tbp]
  \centering
    \includegraphics[width=0.82\linewidth]{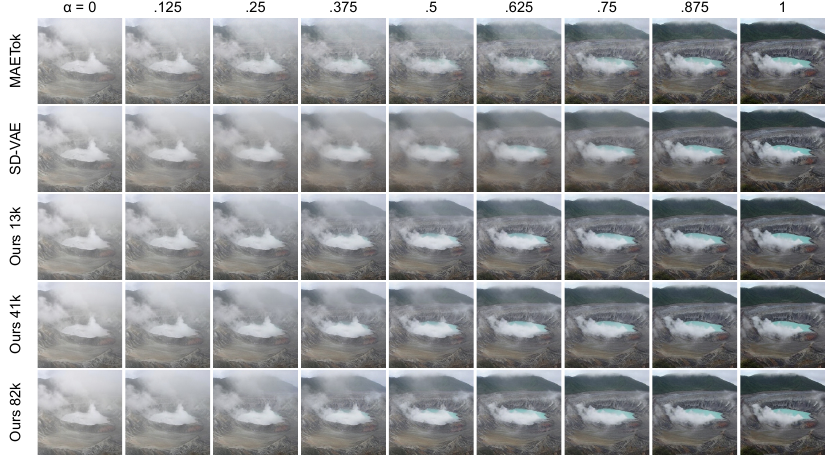}%
  
  \caption{\textbf{Selected interpolation, volcano.}}
  \label{fig:c_pair_03}
\end{figure}

\begin{figure}[!tbp]
  \centering
    \includegraphics[width=0.82\linewidth]{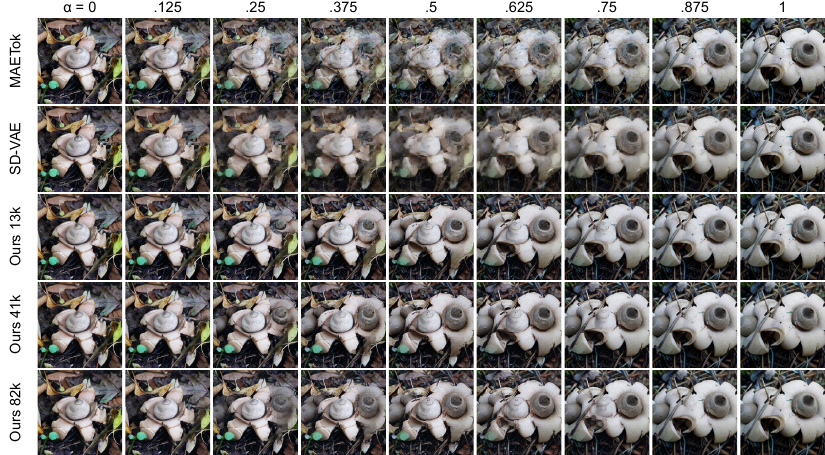}%
  
  \caption{\textbf{Selected interpolation, earthstar, first pair.}}
  \label{fig:c_pair_04}
\end{figure}

\begin{figure}[!tbp]
  \centering
    \includegraphics[width=0.82\linewidth]{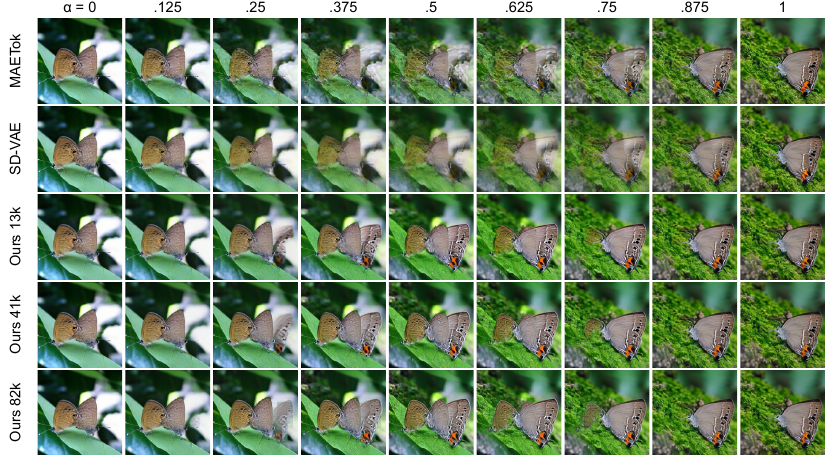}%
  
  \caption{\textbf{Selected interpolation, lycaenid.}}
  \label{fig:c_pair_05}
\end{figure}

\begin{figure}[!tbp]
  \centering
    \includegraphics[width=0.82\linewidth]{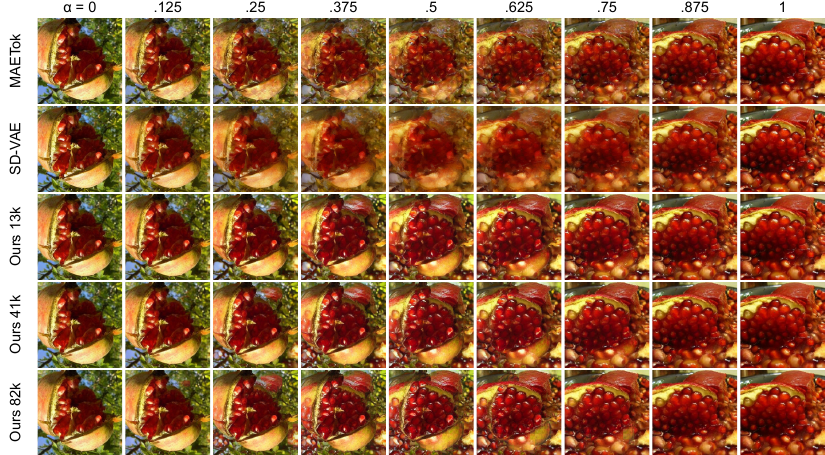}%
  
  \caption{\textbf{Selected interpolation, pomegranate.}}
  \label{fig:c_pair_06}
\end{figure}

\begin{figure}[!tbp]
  \centering
    \includegraphics[width=0.82\linewidth]{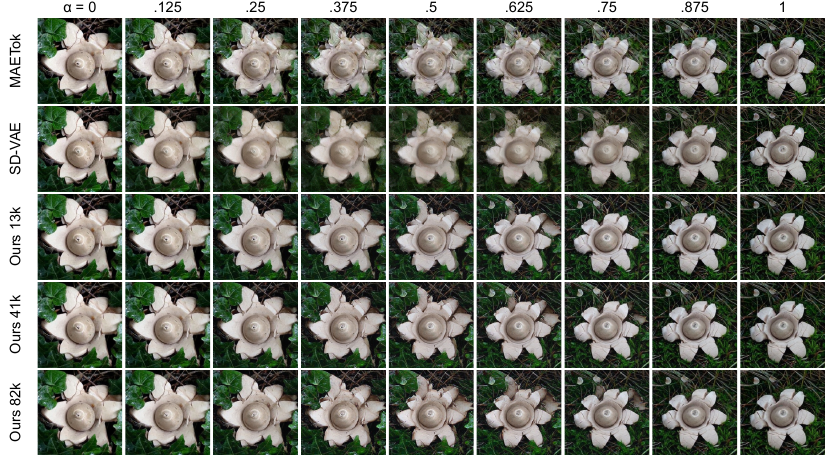}%
  
  \caption{\textbf{Selected interpolation, earthstar, second pair.}}
  \label{fig:c_pair_07}
\end{figure}

\begin{figure}[!tbp]
  \centering
    \includegraphics[width=0.82\linewidth]{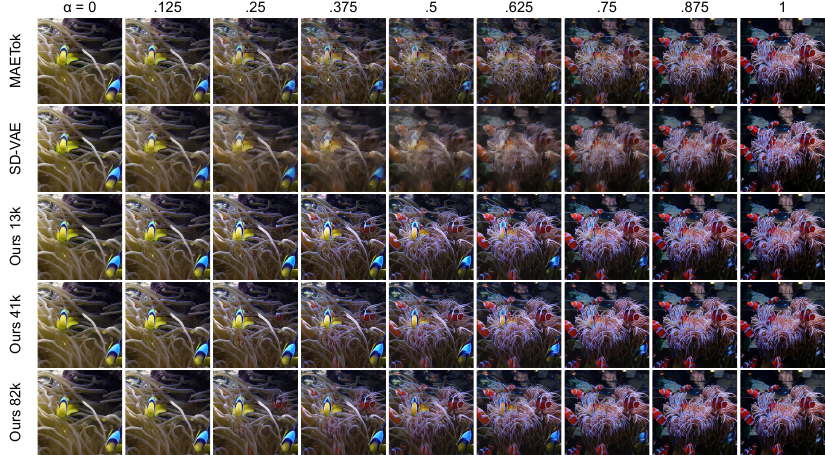}%
  
  \caption{\textbf{Selected interpolation, anemone fish.}}
  \label{fig:c_pair_08}
\end{figure}

\begin{figure}[!tbp]
  \centering
    \includegraphics[width=0.82\linewidth]{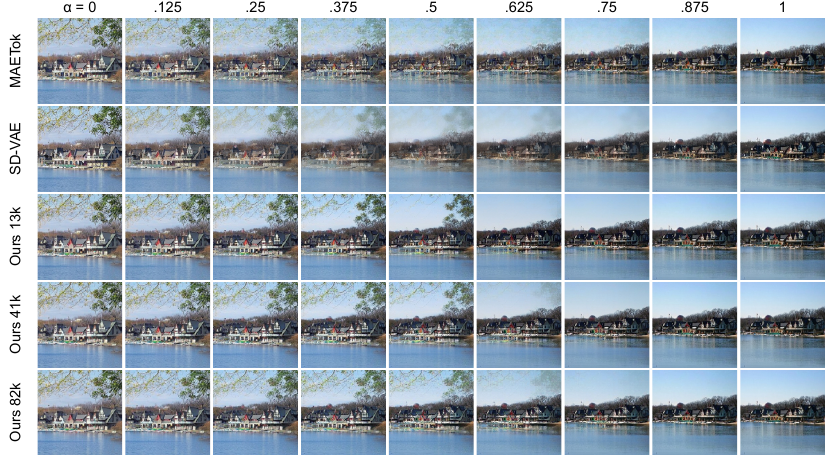}%
  
  \caption{\textbf{Selected interpolation, boathouse.}}
  \label{fig:c_pair_09}
\end{figure}

\begin{figure}[!tbp]
  \centering
    \includegraphics[width=0.82\linewidth]{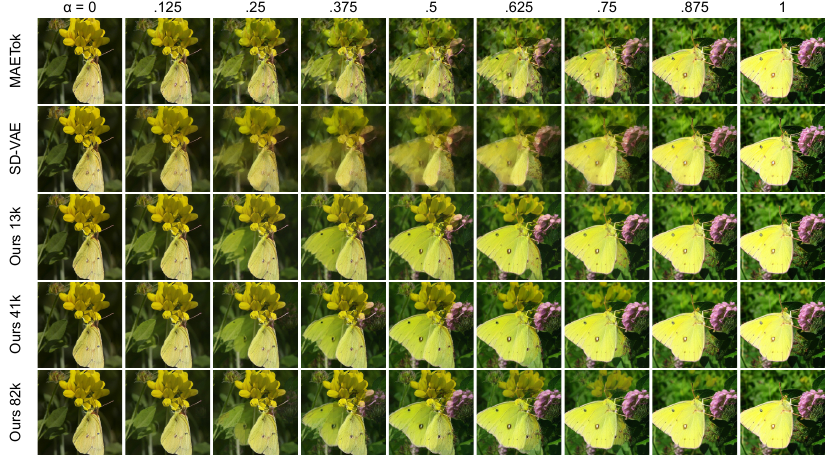}%
  
  \caption{\textbf{Selected interpolation, sulfur butterfly.}}
  \label{fig:c_pair_10}
\end{figure}

\begin{figure}[!tbp]
  \centering
    \includegraphics[width=0.82\linewidth]{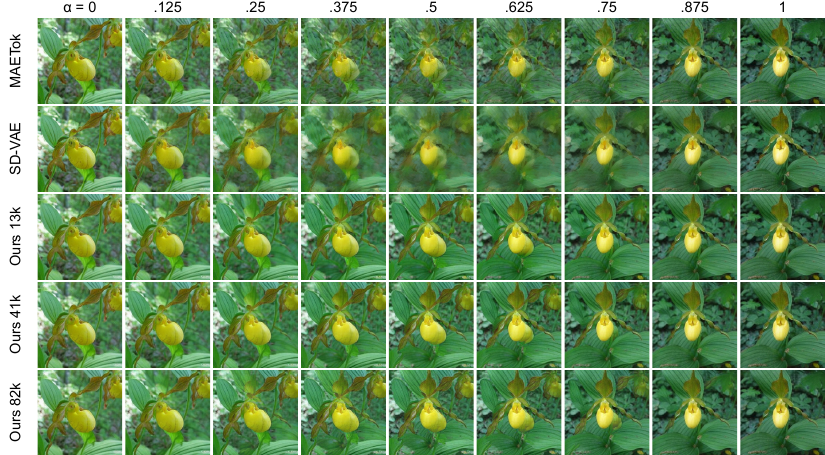}%
  
  \caption{\textbf{Selected interpolation, yellow lady's slipper.}}
  \label{fig:c_pair_11}
\end{figure}

\begin{figure}[!tbp]
  \centering
    \includegraphics[width=0.82\linewidth]{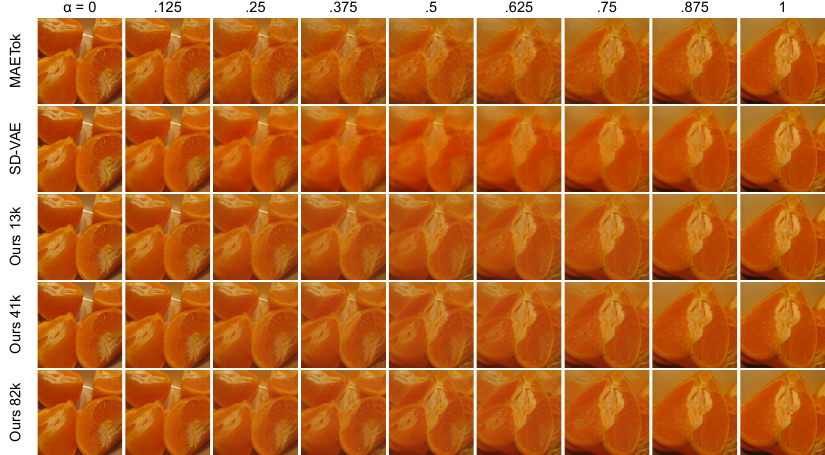}%
  
  \caption{\textbf{Selected interpolation, orange.}}
  \label{fig:c_pair_12}
\end{figure}

\begin{figure}[!tbp]
  \centering
    \includegraphics[width=0.82\linewidth]{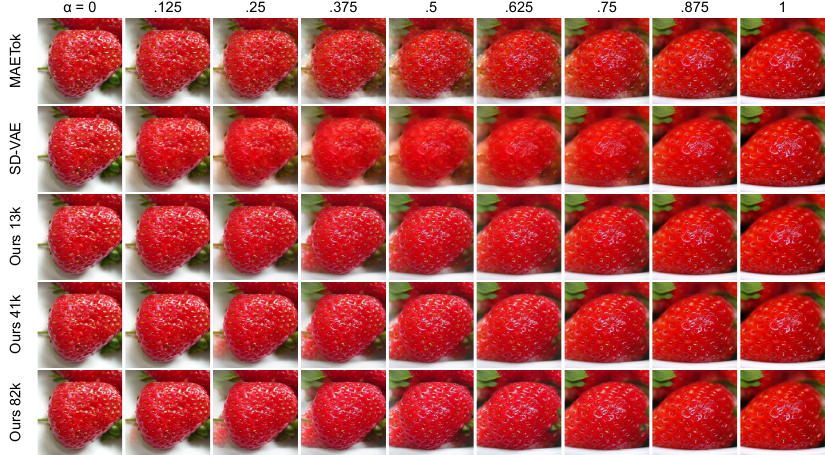}%
  
  \caption{\textbf{Selected interpolation, strawberry.}}
  \label{fig:c_pair_13}
\end{figure}

Figures~\ref{fig:c_pair_01}--\ref{fig:c_pair_13} show thirteen selected pairs at all nine positions, one pair per figure, with the released \maetok, SD-VAE and \ours after 13{,}000, 41{,}000 and 82{,}000 fine-tuning steps from top to bottom.
The sources are ImageNet validation images, which neither \maetok nor \ours trained on.

\begin{figure}[!htb]
  \centering
  \includegraphics[width=0.62\textwidth]{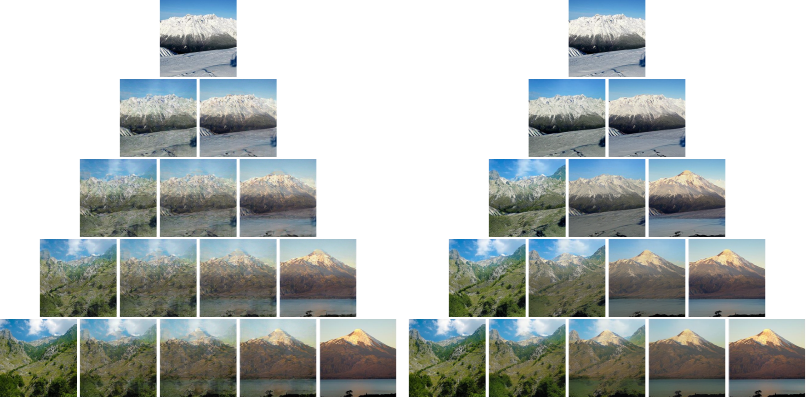}
  \caption{\textbf{Selected three-image interpolation.} \maetok{} left and \ours{} right, with the source images at the corners, mixtures of two neighboring corners along the edges in steps of a quarter and mixtures of all three inside.}
  \label{fig:multi}
\end{figure}

\phantomsection
\label{app:applications}
\label{app:applications:blends}
\textbf{Mixtures of three images.}
Although \ours trains on mixtures of two codes, it also decodes averages of several codes with a lower FID than \maetok, App.~\ref{app:results:kway}.
In Fig.~\ref{fig:multi}, \maetok's decoder decodes mixtures of three mountain images into overlays of the sources, while \ours's decoder produces single mountain scenes.

%% file: appendix/B_additional_results.tex
\FloatBarrier
\section{Additional Results}
\label{app:results}

\subsection{Generation}
\label{app:results:gfid}

\begin{figure}[!htb]
  \centering
  \includegraphics[width=0.60\linewidth]{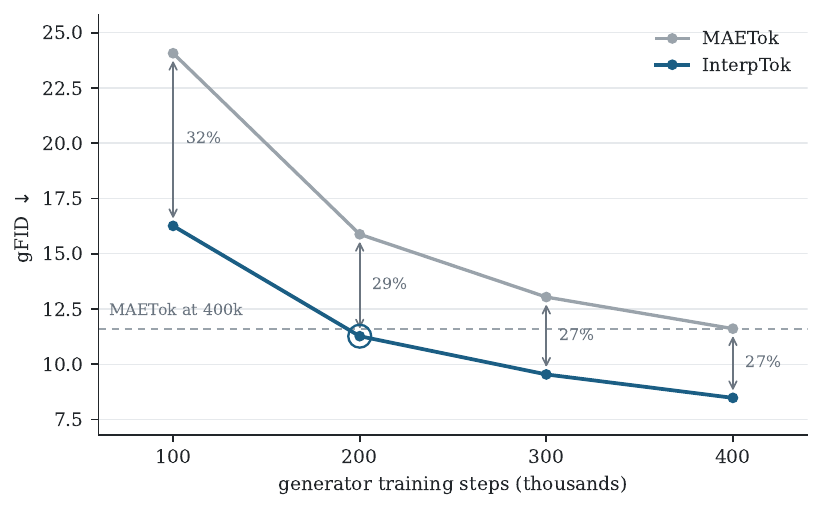}
  \caption{\textbf{Generation across the generator's training.} The percentages are the gap relative to \maetok's gFID at each budget, the dashed rule is \maetok's value at 400{,}000 steps, and the circle marks the first budget at which \ours reaches it.}
  \label{fig:gencurve}
\end{figure}

Every SiT-B/1D result in this paper comes from generators we trained under one recipe, and our \maetok generator is the control against which \ours is compared.
This control reaches a gFID of 11.61, lower than the 13.73 that \citet{xu26ifid} release for a \maetok generator of this size trained for 400{,}000 steps at global batch 128, where ours trains at batch 256.
Table~\ref{tab:gen_all} reports our SiT-B/1D generators~\citep{ma24sit}, trained in the released pipeline of \citet{xu26ifid}, beside published configurations.
Fig.~\ref{fig:gencurve} follows them across the generator's training, Table~\ref{tab:gen_fd} scores the same samples in four feature spaces, and Fig.~\ref{fig:gen_cfg1} shows selected samples.
\ours leads to a lower gFID than \maetok at every generator budget we measured, with the gap widest early and settling near a quarter of \maetok's gFID, Fig.~\ref{fig:gencurve}.
Both fine-tuned checkpoints lead to a lower Fr\'echet distance than \maetok, the 41{,}000-step one in all four feature spaces, including DINOv2, which none of our losses uses, and the 82{,}000-step one in three of four.
SD-VAE reaches a higher PSNR and SSIM and a lower LPIPS than \maetok, Table~\ref{tab:recon}, and its published SiT-B/2 a higher gFID, Table~\ref{tab:gen_all}.

\begin{table}[!htb]
\centering\scriptsize\setlength{\tabcolsep}{4pt}
\caption{\textbf{Generation across generator training budgets, beside the published configurations.} On the left, one SiT-B/1D per row on the named tokenizer's codes, sampled at 250 steps without guidance. On the right, values as published, sampled without guidance, whose generator size, patch size and training budget differ across rows. $n = 50{,}000$ per cell.}
\label{tab:gen_all}
\begin{minipage}[t]{0.42\textwidth}
  \centering\scriptsize\setlength{\tabcolsep}{4pt}
  \begin{tabular}{@{}l r r@{}}
    \toprule
    Tokenizer & generator steps & \gfid\,$\downarrow$ \\
    \midrule
    \maetok{} & 100k & 24.07 \\
    \ours{}   & 100k & 16.26 \\
    \midrule
    \maetok{} & 200k & 15.88 \\
    \ours{}   & 200k & 11.27 \\
    \midrule
    \maetok{} & 300k & 13.04 \\
    \ours{}   & 300k & 9.54 \\
    \midrule
    \maetok{} & 400k & 11.61 \\
    \ours{}   & 400k & 8.48 \\
    \bottomrule
  \end{tabular}
\end{minipage}\hfill
\begin{minipage}[t]{0.54\textwidth}
  \centering\scriptsize\setlength{\tabcolsep}{3pt}
  \begin{tabular}{@{}llrr@{}}
    \toprule
    Tokenizer & Published generator & Steps & \gfid \\
    \midrule
    SD-VAE & SiT-B/2~\citep{ma24sit,bhanded25speedrun} & 400k & 33.0 \\
    \maetok & SiT-L~\citep{chen25maetok} & 400k & 5.69 \\
    \maetok & SiT-XL~\citep{chen25maetok} & 4M & 2.31 \\
    \bottomrule
  \end{tabular}
\end{minipage}
\end{table}

\begin{table}[!htb]
  \centering\scriptsize\setlength{\tabcolsep}{4pt}
  \caption{\textbf{Generated samples in four feature spaces.} Each row is a SiT-B/1D at 400{,}000 steps on the named tokenizer's codes, against the 50{,}000 ImageNet validation images and, for ADM FID, against the ADM reference statistics. Each value is a Fr\'echet distance in the features of SigLIP2, DINOv2, a masked autoencoder (MAE)~\citep{he22mae} or Inception, comparable within a column only.}
  \label{tab:gen_fd}
  \begin{tabular}{@{}l r r r r r@{}}
    \toprule
    Tokenizer & SigLIP2 $\downarrow$ & DINOv2 $\downarrow$ & MAE $\downarrow$ & Inception pool $\downarrow$ & ADM FID $\downarrow$ \\
    \midrule
    \maetok{} & 63.72 & 344.8 & 0.825 & 13.20 & 11.61 \\
    \ours\ (41k) & 59.01 & 309.6 & 0.775 & 10.27 & 8.48 \\
    \ours\ (82k) & 60.73 & 323.1 & 0.851 & 11.69 & 9.57 \\
    \bottomrule
  \end{tabular}
\end{table}

\providecommand{\genclass}[1]{\parbox[b]{\dimexpr\linewidth/6\relax}{\centering\scriptsize #1\strut}}
\begin{figure}[t]
  \centering
  \genclass{typewriter keyboard}\genclass{stove}\genclass{bee}\genclass{badger}\genclass{junco}\genclass{ringlet}\\[1pt]
  \includegraphics[width=\linewidth]{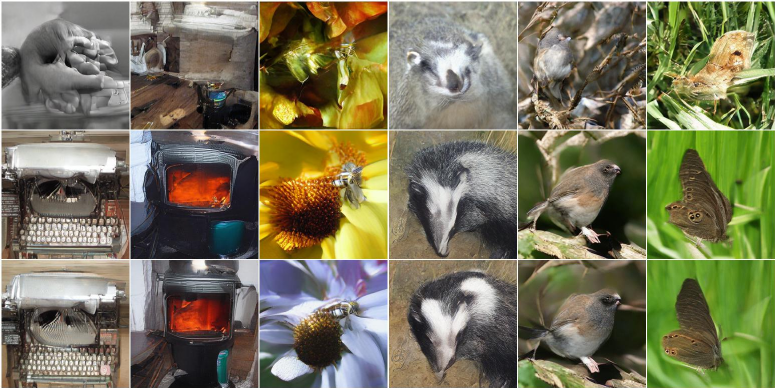}\\[3pt]
  \genclass{Scottish deerhound}\genclass{reflex camera}\genclass{macaw}\genclass{spotted salamander}\genclass{tailed frog}\genclass{meat loaf}\\[1pt]
  \includegraphics[width=\linewidth]{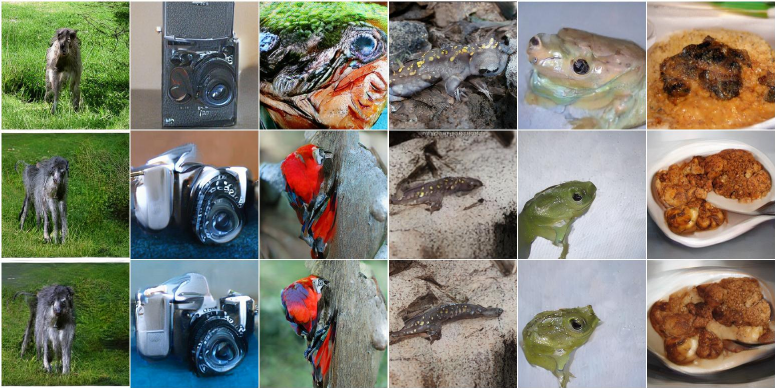}
  \caption{\textbf{Selected samples without guidance.} In each block of three rows, \maetok\ above, \ours\ (41k) in the middle and \ours\ (82k) below. Columns share a class and an initial noise.}
  \label{fig:gen_cfg1}
\end{figure}

\subsectionnobarrier{Decoding a fixed generator's samples}
\label{app:results:swap}

Table~\ref{tab:swap} decodes the same generated codes with \maetok's decoder and with ours.
For our SiT-B on \maetok's codes, our decoder lowers the Fr\'echet distance in all four feature spaces, including DINOv2, which none of our losses uses.
For the released SiT-XL, which we did not train, we compare the two decoders on our own set of samples from that generator, which scores 2.46 with \maetok's decoder where its authors publish 2.31.
In Table~\ref{tab:nfe}, the generator, the starting noise and the sampler are the same within every column, and the gFID gap of our decoder widens with every reduction of the sampling steps.

\begin{table}[!htb]
\centering
\begin{minipage}[t]{0.60\textwidth}
  \centering\scriptsize\setlength{\tabcolsep}{3pt}
  \caption{\textbf{One generator, two decoders.} $n = 50{,}000$ samples per block, decoded by \maetok's decoder and by our decoder, comparable within a column only.}
  \label{tab:swap}
  \begin{tabular}{@{}lrrrrr@{}}
    \toprule
    \multicolumn{6}{@{}l}{\emph{our SiT-B on \maetok's codes}} \\
    Decoder & FID $\downarrow$ & SigLIP2 $\downarrow$ & DINOv2 $\downarrow$ & MAE $\downarrow$ & Inception $\downarrow$ \\
    \midrule
    \maetok & 11.61 & 63.72 & 344.8 & 0.825 & 13.20 \\
    \quad $+$ our decoder & 8.56 & 55.97 & 274.3 & 0.669 & 9.82 \\
    \bottomrule
  \end{tabular}
\end{minipage}\hfill
\begin{minipage}[t]{0.38\textwidth}
  \centering\scriptsize\setlength{\tabcolsep}{3pt}
  \vspace{2.2\baselineskip}
  \begin{tabular}{@{}lr@{}}
    \toprule
    \multicolumn{2}{@{}l}{\emph{the released \maetok SiT-XL~\citep{chen25maetok}}} \\
    Decoder & FID $\downarrow$ \\
    \midrule
    \maetok & 2.46 \\
    \quad $+$ our decoder & 2.26 \\
    \bottomrule
  \end{tabular}
\end{minipage}
\end{table}

\subsectionnobarrier{Image representations}
\label{app:results:probe}

\begin{figure}[!htb]
\centering
\includegraphics[width=\textwidth]{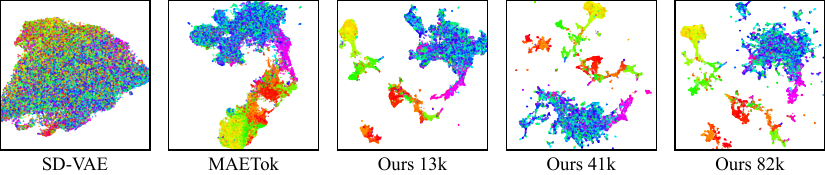}
\caption{\textbf{UMAP projections of image codes.} $n = 50{,}000$ ImageNet validation codes per panel, colored by class.}
\label{fig:umap}
\end{figure}

Table~\ref{tab:mech_latent} gives the linear-probe accuracy on flattened codes under the MAE procedure of Table~\ref{tab:protocol}, and the probe row of Table~\ref{tab:appB_ladder} follows it across the fine-tune.
Under the same procedure, a linear probe predicts the class from SD-VAE's codes far less accurately than from \maetok's.
RAE reports a higher accuracy for its frozen foundation encoder under a different procedure, frozen DINOv2-B linear probing.
Fig.~\ref{fig:umap} projects the validation codes of each tokenizer with the released implementation of UMAP~\citep{mcinnes18umap} at its default settings, with a square window in every panel so that apparent density is comparable.
Its three right-hand panels show three checkpoints of one fine-tuning run.

\subsection{Evaluation during training}
\label{app:results:telemetry}

Table~\ref{tab:appB_ladder} follows four checkpoints of one fine-tuning run. Its two blocks of pairs differ in pairing, in straight or spherical path and in reference images.
The random-pair midpoint FID falls more than sixfold from the released model to 41{,}000 steps and changes little afterwards.
The reconstruction FID keeps falling after 13{,}000 steps, and the linear probe keeps improving.
Both the 41{,}000-step and the 82{,}000-step checkpoints lead to a lower gFID than \maetok.
We report 41{,}000 steps, since the generator on these codes reaches the lowest gFID, Table~\ref{tab:gen}, and every variant of Table~\ref{tab:abl} that trains encoder and decoder is matched there.
The 13{,}000-step checkpoint has the lowest \ifid, but a generator on its codes reaches a higher gFID than one on the 41{,}000-step codes, Table~\ref{tab:gen}.
We read this as generation depending on more than the decodability that \ifid measures, in line with \citet{zhong26diffusability}.

\begin{table}[!htb]
\centering
\input{tables/B5_generation.tex}\hfill
\begin{minipage}[t]{0.64\textwidth}
  \centering\scriptsize\setlength{\tabcolsep}{3pt}
  \caption{\textbf{Checkpoints of the fine-tune}, $n = 50{,}000$ per FID.}
  \label{tab:appB_ladder}
  \begin{tabular}{@{}lrrrrr@{}}
    \toprule
    Cross-class pairs & \maetok & 13k & 41k & 48k & 82k \\
    \midrule
    midpoint FID\,$\downarrow$ & 87.34 & 16.06 & 13.88 & 13.86 & 14.92 \\
    reconstruction FID\,$\downarrow$ & 0.575 & 0.581 & 0.574 & 0.561 & 0.553 \\
    \bottomrule
  \end{tabular}
  \par\medskip
  \begin{tabular}{@{}lrrrr@{}}
    \toprule
    Nearest-neighbor pairs & \maetok & 13k & 41k & 82k \\
    \midrule
    \ifid\,$\downarrow$ & 14.42 & 3.97 & 4.30 & 4.47 \\
    \rfid\,$\downarrow$ & 0.582 & 0.601 & 0.597 & 0.574 \\
    \bottomrule
  \end{tabular}
  \par\medskip
  \begin{tabular}{@{}lrrr@{}}
    \toprule
    Codes & \maetok & 41k & 82k \\
    \midrule
    probe\,$\uparrow$ & 73.0 & 77.9 & 78.4 \\
    \gfid\,$\downarrow$ & 11.61 & 8.48 & 9.57 \\
    \bottomrule
  \end{tabular}
\end{minipage}
\end{table}

\subsection{Reconstruction on ImageNet and MS-COCO}
\label{app:results:ood}

The main results pair each code with one of its nearest training codes and decode their midpoint.
Here, we vary the pairing, the position on the line, the number of mixed codes and the dataset.
\ours has higher interpolation quality than \maetok under every pairing, number of codes and dataset we measure, with the largest advantage near the middle of the line.

\Needspace{15\baselineskip}
\begin{wraptable}{r}{0.55\textwidth}
\vspace{-\intextsep}
  \centering\scriptsize
  \caption{\textbf{MS-COCO reconstruction and decoded midpoints.} FID against the 50{,}000 train2017 sources, $n = 50{,}000$ per cell.}
  \label{tab:coco_fid}
  \begin{tabular}{@{}l r r@{}}
    \toprule
    Tokenizer & Reconstruction $\downarrow$ & Midpoint $\downarrow$ \\
    \midrule
    SD-VAE & \textbf{0.801} & 135.89 \\
    \maetok & 0.863 & 85.21 \\
    \quad w/ interpolation & 0.865 & \textbf{9.23} \\
    \midrule
    \ours & 1.013 & 13.61 \\
    \bottomrule
  \end{tabular}
\end{wraptable}

Tables~\ref{tab:recon}, \ref{tab:coco_fid} and~\ref{tab:coco_fd} report reconstruction on ImageNet and MS-COCO and the decoded MS-COCO midpoints in three feature spaces.
On MS-COCO, where neither tokenizer is trained, \ours's midpoint FID is six times lower than \maetok's.
The variant trained without the regularization losses has an even lower midpoint FID.
Among these tokenizers, SD-VAE has the lowest reconstruction FID and the highest midpoint FID.

\Needspace{13\baselineskip}
\begin{wraptable}{r}{0.55\textwidth}
\vspace{-\intextsep}
  \centering\scriptsize
  \caption{\textbf{Decoded MS-COCO midpoints in three feature spaces}, $n = 50{,}000$ midpoints of random pairs of different train2017 images, against the COCO sources. Values are comparable within a column only.}
  \label{tab:coco_fd}
  \begin{tabular}{l r r r}
    \toprule
    Tokenizer & SigLIP2 $\downarrow$ & MAE $\downarrow$ & DINOv2 $\downarrow$ \\
    \midrule
    \maetok{} & 85.8 & 2.03 & 953 \\
    \ours\ (41k) & 34.6 & 0.52 & 315 \\
    \ours\ (82k) & 36.4 & 0.70 & 311 \\
    \bottomrule
  \end{tabular}
\end{wraptable}

In the features of DINOv2, which \maetok's auxiliary decoders regress and none of our losses uses, \ours's midpoint Fr\'echet distance is still three times lower.
The 41{,}000-step checkpoint reaches a lower Fr\'echet distance than the 82{,}000-step one in the SigLIP2 and MAE features, which the representation Fr\'echet loss uses, and both reach about the same in DINOv2.

\input{tables/T2_recon.tex}

\subsection{Mixtures of several image codes}
\label{app:results:kway}
\label{app:kway}

Table~\ref{tab:kway} decodes equal-weight averages of several codes and compares them with the originals of each corpus.
Across classes, the FID of \ours's decoded averages of two codes is six times lower than \maetok's, and for averages of ten codes still slightly lower.
On same-class pairs, the Inception Score~\citep{salimans16is} of \ours's decoded mixtures is about the same as that of its reconstructions.

\input{tables/B2_kway.tex}

\subsectionnobarrier{Decoded paths}
\label{app:results:paths}

A path can look natural at every position and still fade slowly from one image into the other, or change abruptly near the middle.
Table~\ref{tab:ppl} distinguishes these two cases from a path that changes steadily, and Table~\ref{tab:path} measures how natural the decoded images look along the path.
Table~\ref{tab:ppl} gives the perceptual path length (PPL) after \citet{karras19stylegan}, here the \lpips between consecutive decoded images summed over nine positions.
Its share is the largest step between consecutive decoded images as a share of the path length. Its $d_{\mathrm{mid}}$ is the distance from the decoded midpoint to a source reconstruction as a share of the distance between the two source reconstructions, zero for a path that copies a source at the middle.
Its ordered column is the share of pairs whose decoded images move away from the first source and toward the second at every one of the nine positions.
Table~\ref{tab:path} compares the images decoded along the path with held-out real images.
CMMD is the distance of \citet{jayasumana24cmmd} in CLIP~\citep{radford21clip} features, in thousandths.
An autoencoder crossfade blends a tokenizer's two reconstructions, and a pixel crossfade blends the two source images.
\ours's paths reach a lower CMMD than those of \maetok and SD-VAE and than every crossfade, Table~\ref{tab:path}, and nearly all of them move in order from the first source to the second, Table~\ref{tab:ppl}.

\input{tables/B8_ppl.tex}
\input{tables/B3_path.tex}

\subsectionnobarrier{Interpolation position and geometry}
\label{app:results:position}

We also decode other positions on the line and a spherical path, Table~\ref{tab:geom}.
\ours's advantage over \maetok is largest near the middle of the line, where decoding is hardest.
On a spherical path, which \citet{white16sampling} recommends for Gaussian latent spaces, \ours's midpoint FID is even lower than on the straight line.
\citet{arvanitidis18oddity} also show that the straight line is not the shortest path in a curved latent space.
Diffusion methods that interpolate between two images or two text prompts change the generator or its sampling, not the tokenizer~\citep{wang23interpolating,zhang24diffmorpher,yang24impus,zheng24noisediffusion,he24aid,guo24smoothdiffusion}.

\begin{table}[!htb]
  \centering\small
  \caption{\textbf{Position along the path and interpolation geometry}, FID on $n = 50{,}000$ decoded images per cell. In panel b, near pairs each code with its nearest validation code in the latent space of the model, as in Table~\ref{tab:evalvar}.}
  \label{tab:geom}
  \scriptsize\setlength{\tabcolsep}{3pt}
  \begin{minipage}[t]{0.48\textwidth}\centering
  \textit{(a) Position beyond the midpoint}\par\smallskip
  \begin{tabular}{@{}lrr@{}}
    \toprule
    Position & \maetok & \ours\ (41k) \\
    \midrule
    $\alpha=0.625$ & 10.17 & 5.35 \\
    $\alpha=0.75$  & 6.12  & 5.62 \\
    $\alpha=0.875$ & 4.76  & 5.25 \\
    \bottomrule
  \end{tabular}
  \end{minipage}\hfill
  \begin{minipage}[t]{0.48\textwidth}\centering
  \textit{(b) Pairing and path}\par\smallskip
  \begin{tabular}{@{}lrrr@{}}
    \toprule
    Pairing and path & \maetok & \ours\ (41k) & \ours\ (82k) \\
    \midrule
    Near, straight    & 17.28 & 5.58  & 5.82  \\
    Near, spherical   & 17.15 & 4.27  & 4.38  \\
    \bottomrule
  \end{tabular}
  \end{minipage}
\end{table}

%% file: tables/B5_generation.tex
\begin{minipage}[t]{0.32\textwidth}
\caption{\textbf{Generation from tokenizer codes.} Matched SiT-B/1D generators at 250 sampling steps without guidance, $n = 50{,}000$.}
\label{tab:gen}
\centering\scriptsize\setlength{\tabcolsep}{4pt}
\begin{tabular}{@{}lr@{}}
\toprule
Tokenizer & \gfid\,$\downarrow$ \\
\midrule
\maetok & 11.61 \\
\ours\ (13k) & 8.94 \\
\ours\ (41k) & \textbf{8.48} \\
\ours\ (82k) & 9.57 \\
\bottomrule
\end{tabular}
\end{minipage}

%% file: tables/T2_recon.tex
\begin{table}[!htb]
\caption{\textbf{Reconstruction on ImageNet and MS-COCO.} The ImageNet validation set and MS-COCO train2017, $n = 50{,}000$ per cell. $^{\dagger}$Quoted from \citet{chen25maetok}.}
\label{tab:recon}
\centering\footnotesize\setlength{\tabcolsep}{2pt}
\resizebox{\linewidth}{!}{%
\begin{tabular}{@{}lccrrrrrrrr@{}}
\toprule
& & & \multicolumn{4}{c}{ImageNet} & \multicolumn{4}{c}{MS-COCO} \\
\cmidrule(lr){4-7}\cmidrule(lr){8-11}
Tokenizer & Params (M) & Tokens & \rfid\,$\downarrow$ & PSNR\,$\uparrow$ & SSIM\,$\uparrow$ & \lpips\,$\downarrow$ & \rfid\,$\downarrow$ & PSNR\,$\uparrow$ & SSIM\,$\uparrow$ & \lpips\,$\downarrow$ \\
\midrule
SD-VAE & 84$^{\dagger}$ & 1024 & 0.62$^{\dagger}$ & \textbf{25.00} & \textbf{0.805} & \textbf{0.0630} & \textbf{0.801} & \textbf{24.76} & \textbf{0.817} & \textbf{0.0609} \\
\maetok & 174 & 128 & \textbf{0.582} & 23.55 & 0.760 & 0.0917 & 0.863 & 23.25 & 0.770 & 0.0913 \\
\midrule
\ours & 217 & 128 & 0.597 & 24.02 & 0.775 & 0.0819 & 1.013 & 23.65 & 0.784 & 0.0817 \\
\bottomrule
\end{tabular}}
\end{table}

%% file: tables/B2_kway.tex
\begin{table}[!htb]
\caption{\textbf{Image quality of decoded mixtures.} Equal-weight code mixtures on ImageNet and MS-COCO against each corpus's own 50{,}000 originals, $n = 50{,}000$ tuples per cell. Adjacent codes are those of consecutive validation images of one class. sFID, precision and recall are computed with the ADM evaluator~\citep{dhariwal21adm}, precision and recall after \citet{kynkaanniemi19precision}.}
\label{tab:kway}
\centering\footnotesize\setlength{\tabcolsep}{3pt}
\begin{tabular}{@{}lrrrrrrrrrr@{}}
\toprule
& \multicolumn{9}{c}{ImageNet} & \multicolumn{1}{c}{MS-COCO} \\
\cmidrule(lr){2-10}\cmidrule(lr){11-11}
& 1 code & \multicolumn{4}{c}{adjacent codes, same class} & \multicolumn{4}{c}{codes of random classes} & 2, random \\
\cmidrule(lr){3-6}\cmidrule(lr){7-10}
& & $k{=}2$ & $3$ & $5$ & $10$ & $k{=}2$ & $3$ & $5$ & $10$ & \\
\midrule
\multicolumn{11}{@{}l}{\textit{FID\,$\downarrow$}} \\
\maetok & 0.57 & 35.12 & 72.80 & 121.02 & 181.21 & 87.49 & 143.46 & 186.22 & 238.52 & 85.21 \\
\ours & 0.57 & \textbf{3.79} & \textbf{6.71} & \textbf{18.59} & \textbf{76.23} & \textbf{13.78} & \textbf{40.81} & \textbf{124.85} & \textbf{219.07} & \textbf{13.61} \\
\midrule
\multicolumn{11}{@{}l}{\textit{Inception Score\,$\uparrow$}} \\
\maetok & 138.1 & 47.7 & 24.5 & 11.4 & 4.9 & 15.8 & 8.0 & 4.5 & 2.4 & 10.6 \\
\ours & \textbf{140.1} & \textbf{139.4} & \textbf{131.2} & \textbf{92.0} & \textbf{24.9} & \textbf{58.0} & \textbf{30.4} & \textbf{11.4} & \textbf{2.7} & \textbf{20.7} \\
\midrule
\multicolumn{11}{@{}l}{\textit{sFID\,$\downarrow$}} \\
\maetok & 3.73 & 20.13 & 36.26 & 66.33 & 130.75 & 23.11 & 41.34 & 77.95 & 160.41 & 27.39 \\
\ours & \textbf{3.25} & \textbf{10.05} & \textbf{14.57} & \textbf{19.83} & \textbf{47.17} & \textbf{13.92} & \textbf{21.81} & \textbf{30.44} & \textbf{85.44} & \textbf{15.65} \\
\midrule
\multicolumn{11}{@{}l}{\textit{Precision\,$\uparrow$}} \\
\maetok & 0.999 & 0.442 & 0.294 & 0.228 & 0.130 & 0.187 & 0.141 & 0.134 & 0.048 & 0.189 \\
\ours & \textbf{1.000} & \textbf{0.917} & \textbf{0.859} & \textbf{0.664} & \textbf{0.350} & \textbf{0.510} & \textbf{0.386} & \textbf{0.248} & \textbf{0.273} & \textbf{0.535} \\
\midrule
\multicolumn{11}{@{}l}{\textit{Recall\,$\uparrow$}} \\
\maetok & 0.999 & 0.562 & 0.339 & 0.127 & 0.008 & 0.572 & 0.177 & 0.014 & 0.000 & 0.397 \\
\ours & 0.999 & \textbf{0.612} & \textbf{0.350} & \textbf{0.157} & \textbf{0.047} & \textbf{0.787} & \textbf{0.636} & \textbf{0.323} & \textbf{0.014} & \textbf{0.697} \\
\bottomrule
\end{tabular}
\end{table}

%% file: tables/B8_ppl.tex
\begin{table}[!htb]
\caption{\textbf{Movement along decoded paths.} Path length in the spirit of \citet{karras19stylegan} with step shares and source distances, over nine equally spaced positions of the straight path, $n = 5{,}000$ per pair set.}
\label{tab:ppl}
\centering\footnotesize\setlength{\tabcolsep}{4pt}
\begin{tabular}{@{}lrrrrrrrr@{}}
\toprule
& \multicolumn{4}{c}{near pairs} & \multicolumn{4}{c}{random pairs} \\
\cmidrule(lr){2-5}\cmidrule(lr){6-9}
Path & PPL & share & $d_{\mathrm{mid}}$ & ordered\,$\uparrow$ & PPL & share & $d_{\mathrm{mid}}$ & ordered\,$\uparrow$ \\
\midrule
SD-VAE & 0.796 & 0.174 & 0.739 & 61.4\% & 0.899 & 0.186 & 0.722 & 58.4\% \\
\maetok & 0.665 & 0.167 & 0.623 & 99.8\% & 0.799 & 0.172 & 0.636 & 99.4\% \\
\ours & 0.779 & 0.197 & 0.603 & 99.1\% & 1.056 & 0.203 & 0.628 & 94.8\% \\
\bottomrule
\end{tabular}
\par\medskip
\begin{tabular}{@{}lrrrr@{}}
\toprule
& \multicolumn{2}{c}{near pairs} & \multicolumn{2}{c}{random pairs} \\
\cmidrule(lr){2-3}\cmidrule(lr){4-5}
Reference path & PPL & share & PPL & share \\
\midrule
autoencoder crossfade of \ours & 0.433 & 0.198 & 0.530 & 0.215 \\
endpoint copy, $d_{\mathrm{mid}} = 0$ & 0.671 & 1.0 & 0.768 & 1.0 \\
\bottomrule
\end{tabular}
\end{table}

%% file: tables/B3_path.tex
\begin{table}[t]
\caption{\textbf{Image quality along paths.} CMMD in thousandths of the images decoded along the straight path between the codes of two images of different classes, against held-out real images, $n = 25{,}000$ paths per row. The reconstruction ceiling scores each tokenizer's reconstructions of real images.}
\label{tab:path}
\centering\scriptsize\setlength{\tabcolsep}{3pt}
\begin{minipage}[t]{0.48\textwidth}\centering
\textit{(a) Paths between two images}\par\smallskip
\begin{tabular}{@{}lr@{}}
\toprule
Path & CMMD $\downarrow$ \\
\midrule
\multicolumn{2}{@{}l}{\textit{\ours}} \\
\quad line in the latent space & \textbf{171.3} \\
\quad autoencoder crossfade & 462.4 \\
\multicolumn{2}{@{}l}{\textit{\maetok}} \\
\quad line in the latent space & 508.5 \\
\quad autoencoder crossfade & 493.4 \\
\multicolumn{2}{@{}l}{\textit{SD-VAE}} \\
\quad line in the latent space & 857.3 \\
\quad autoencoder crossfade & 598.4 \\
\midrule
pixel crossfade & 368.3 \\
\bottomrule
\end{tabular}
\end{minipage}\hfill
\begin{minipage}[t]{0.48\textwidth}\centering
\textit{(b) Reconstruction ceiling}\par\smallskip
\begin{tabular}{@{}lr@{}}
\toprule
Tokenizer & CMMD $\downarrow$ \\
\midrule
\ours & 57.7 \\
\maetok & 72.9 \\
SD-VAE & 98.9 \\
\bottomrule
\end{tabular}
\end{minipage}
\end{table}

%% file: appendix/I_mechanism.tex
\subsection{Why Generation Quality Changes with the Sampling Steps, the Sampler and the Training Objective}
\label{app:mechanism}

\textbf{Observation.}
With the stochastic sampler of Table~\ref{tab:nfe}, \ours's generator reaches a lower gFID with 64 or 32 sampling steps than with 250, while every other generator worsens with every reduction of the steps, Table~\ref{tab:mech_fid}.
These include the generator on \maetok's codes with \maetok's decoder, with our decoder and with a decoder that adds only the interpolation discriminator to \maetok's losses, the generators of the two variants trained without interpolation and without the regularization losses, and the released SiT-XL with either decoder.

\begin{table}[!htbp]
\centering\scriptsize
\caption{\textbf{Generation quality with the stochastic and the deterministic sampler.} \gfid of the class-conditional SiT-B generators of Sec.~\ref{sec:setup} and of the released SiT-XL against the ADM statistics of ImageNet 256 without guidance, $n = 50{,}000$ per cell. Without interp.\ and without reg.\ are the variants of Table~\ref{tab:abl} trained without interpolation and without the regularization losses, the rows w/ reg.\ and w/ interpolation there, and $+$ marks a decoder placed on the codes of the row above.}
\label{tab:mech_fid}
\begin{minipage}[t]{0.56\textwidth}\centering\scriptsize\setlength{\tabcolsep}{3pt}
\textit{(a) stochastic sampler}\par\smallskip
\begin{tabular}{@{}lrrrr@{}}
\toprule
\gfid\,$\downarrow$ & 250 & 64 & 32 & 16 \\
\midrule
\multicolumn{5}{@{}l}{\textit{SiT-B on \maetok's codes, one generator}} \\
\maetok & 11.61 & 12.33 & 13.88 & 20.56 \\
\quad $+$ our decoder & 8.56 & 8.99 & 9.92 & 15.33 \\
\quad $+$ \maetok's losses w/ interp.\ discriminator & 9.71 & 10.16 & 11.26 & 16.92 \\
\quad $+$ \maetok's losses & 10.80 & 11.50 & 13.05 & 20.02 \\
\midrule
\multicolumn{5}{@{}l}{\textit{SiT-B on each tokenizer's own codes}} \\
\ours & 8.48 & 8.09 & 8.11 & 12.48 \\
without reg. & 18.90 & 19.68 & 20.99 & 27.79 \\
\midrule
\multicolumn{5}{@{}l}{\textit{released SiT-XL on \maetok's codes, one generator}} \\
\maetok & 2.46 & 2.78 & 3.46 & 7.52 \\
\quad $+$ our decoder & 2.26 & 2.45 & 2.79 & 5.80 \\
\bottomrule
\end{tabular}
\par\medskip
\begin{tabular}{@{}lrrr@{}}
\toprule
\gfid\,$\downarrow$, stochastic sampler & 250 & 64 & 32 \\
\midrule
without interp., on its own codes & 21.15 & 21.53 & 22.98 \\
\bottomrule
\end{tabular}
\end{minipage}\hfill
\begin{minipage}[t]{0.42\textwidth}\centering\scriptsize\setlength{\tabcolsep}{3pt}
\textit{(b) deterministic sampler}\par\smallskip
\begin{tabular}{@{}lrrrrr@{}}
\toprule
\gfid\,$\downarrow$ & 250 & 64 & 32 & 16 & 8 \\
\midrule
\maetok & 15.74 & 15.43 & 15.52 & 16.01 & 19.41 \\
\quad $+$ our decoder & 13.25 & -- & 11.47 & 9.82 & 7.50 \\
\ours & 16.17 & 14.02 & 11.49 & 7.93 & 6.38 \\
\bottomrule
\end{tabular}
\par\medskip
\begin{tabular}{@{}lrrr@{}}
\toprule
\gfid\,$\downarrow$, deterministic sampler & 250 & 32 & 16 \\
\midrule
without interp. & 29.14 & 24.98 & 22.64 \\
without reg. & 23.09 & 20.68 & 18.05 \\
\bottomrule
\end{tabular}
\end{minipage}
\end{table}

\textbf{Hypothesis.}
Generated samples of a diffusion model tend to lie between the modes of its training data~\citep{aithal24hallucination}.
The modes are the clusters of similar samples in the training data.
In each sampling step, the model predicts the clean code from the noisy code as it currently looks, and the sampler moves the noisy code part of the way toward this prediction.
The prediction that minimizes the denoising loss on the training data is an average of the training codes, each weighted by how well it matches the noisy code~\citep{xu26ifid}.
When codes of several modes match, this average lies between these modes.
When the noise is strong, the noisy code carries little information about the clean code, and the prediction approaches the average of all training codes~\citep{karras22edm}.
A conditioned generator also receives a condition, here the class label.
At the strongest noise, the noisy code reveals nothing about the clean code, so the class label is the only cue, and the prediction approaches the average of the training codes of that class, which we call the class average.
As the noise falls, patterns appear in the noisy code, coarse ones first and fine ones last~\citep{skorokhodov25diffusability}, and the prediction moves away from the class average toward the average of the training codes that match these patterns.
The lower the noise, the more the patterns steer the prediction.
With fewer sampling steps, each step is larger, and the generator has fewer steps in which to follow the patterns that emerge in the code. It moves in larger steps toward the class average of its label and ends up closer to that average before the patterns can steer it away.
Whether this helps depends on the tokenizer, since the class average is itself a code between images, the encoder decides which codes lie around it, and the decoder determines how such a code is decoded into pixel space.

\textbf{Stochastic and deterministic sampler.} For the reported gFID values in the paper, including the results with fewer steps, Table~\ref{tab:nfe}, we use the stochastic sampler with which SiT~\citep{ma24sit} reports its gFID, which adds noise after each step. It is the standard choice for SiT generators and usually gives better generation results than a deterministic sampler, which adds no new noise after each step. At 250 steps, the deterministic sampler gives a worse gFID than the stochastic sampler for every generator we test, Table~\ref{tab:mech_fid}. The reasoning in the literature for why the stochastic sampler is better is that the denoiser never predicts perfectly and every finite step adds a small step error~\citep{karras22edm,song21sde,xu23restart}. Without added noise, these small errors build up more from step to step and the generation drifts further, while the added noise reduces this drift. We interpret this error as the model thinking it sees patterns in the code which it also saw during training and moving toward the average of this kind of pattern in the training samples. However, this pattern depends on the random noise and therefore might not resemble a pattern that suits the kind of image we want to generate, for example an image of a specific class. In this reading, the added noise makes the generator diverge later toward some arbitrary pattern, while with the deterministic sampler it diverges sooner. The price of adding noise is that the stochastic sampler needs more steps than a deterministic sampler, since every step also has to remove the noise added after the previous step, which makes the path less smooth~\citep{xu23restart}. With the stochastic sampler, we would therefore expect every model to get worse with fewer steps. In Table~\ref{tab:nfe}, every model except \ours does get worse, while \ours improves down to 32 steps. The following tests examine why \ours behaves differently.

\textbf{Fewer steps.} To isolate the mechanism of the generative model, how it behaves during denoising, we use the deterministic sampler for the following tests. First, we redo the generation with different numbers of steps with the deterministic sampler, to see how the models behave with fewer steps when no noise is added, Table~\ref{tab:mech_fid}. At 250 steps, all models, meaning \maetok, \ours, \maetok's codes with our decoder and the two variants trained without interpolation and without the regularization losses, perform worse than with the stochastic sampler. \maetok's codes with our decoder perform best, while \ours performs slightly worse than \maetok. However, with fewer sampling steps, the gFID of \maetok changes by less than two percent down to 16 steps and only gets worse at 8 steps, while \maetok's codes with our decoder improve from 13.25 to 7.50 and \ours from 16.17 to 6.38. The two variants also improve, by about a fifth at 16 steps, but from a much worse gFID at 250 steps. So without the added noise, the gain of \ours from fewer steps, which we observed with the stochastic sampler, becomes even larger. We hypothesize that without added noise after each step, the deterministic sampler makes the generator pick up on patterns earlier than with the stochastic sampler, which steers the generation further away from the class condition toward similar patterns in codes it saw during training. By our hypothesis, fewer steps make each step bigger, so that the generations end closer to the class average. We estimate the class average from the codes of the 50 ImageNet validation images of each class. To test whether fewer steps bring the generated codes closer to their class average, we measure for every generated code, about 50 per class, its distance to the average code of its class, at 250, 32, 16 and 8 steps, Table~\ref{tab:mech_dist}. With the deterministic sampler, the generated codes of all models are 15 to 18\% closer to their class average at 8 steps than at 250 steps. Part of this is a general move toward the middle of the latent space, since the codes also come closer to the nearest average of another class, by 13\% for \maetok and 10\% for \ours. The distance to their own class average, however, shrinks more than the distance to the nearest other class average, more so for \ours, and the share of codes whose nearest class average is their own rises from 71\% to 87\% for \ours and from 38\% to 48\% for \maetok. With the stochastic sampler, which we measured for \ours and \maetok, the codes also come closer to their class average, but less so, by 13\% for \ours and 4\% for \maetok. We read this as support for the first part of our explanation, that fewer and therefore bigger steps make the generation land closer to its class average.

\begin{table}[!htbp]
\centering\scriptsize\setlength{\tabcolsep}{4pt}
\caption{\textbf{Finished codes and class averages with fewer steps.} Distance of the finished generated codes to the average code of their own class and to the nearest average code of another class, relative to 250 steps, and the share of codes whose nearest class average is their own. 50{,}048 generated codes per generator, with the deterministic sampler unless the row says stochastic. Without interp.\ and without reg.\ are the rows w/ reg.\ and w/ interpolation of Table~\ref{tab:abl}.}
\label{tab:mech_dist}
\begin{tabular}{@{}p{2.8cm}rrrr@{}}
\toprule
Sampling steps & 250 & 32 & 16 & 8 \\
\midrule
\multicolumn{5}{@{}p{5.8cm}@{}}{\textit{a. distance to the average code of its own class, relative to 250 steps}} \\
\maetok & 1 & 0.97 & 0.93 & 0.85 \\
\ours & 1 & 0.96 & 0.91 & 0.82 \\
without interp. & 1 & 0.96 & 0.91 & 0.82 \\
without reg. & 1 & 0.96 & 0.91 & 0.83 \\
\maetok, stochastic & 1 & 0.98 & 0.96 & 0.96 \\
\ours, stochastic & 1 & 0.97 & 0.93 & 0.87 \\
\midrule
\multicolumn{5}{@{}p{5.8cm}@{}}{\textit{b. distance to the nearest average code of another class, relative to 250 steps}} \\
\maetok & 1 & 0.97 & 0.93 & 0.87 \\
\ours & 1 & 0.98 & 0.95 & 0.90 \\
without interp. & 1 & 0.98 & 0.95 & 0.91 \\
without reg. & 1 & 0.96 & 0.92 & 0.84 \\
\maetok, stochastic & 1 & 0.98 & 0.97 & 0.97 \\
\ours, stochastic & 1 & 0.98 & 0.97 & 0.94 \\
\midrule
\multicolumn{5}{@{}p{5.8cm}@{}}{\textit{c. codes whose nearest class average is their own, percent}} \\
\maetok & 38 & 41 & 43 & 48 \\
\ours & 71 & 75 & 79 & 87 \\
without interp. & 72 & 77 & 82 & 89 \\
without reg. & 25 & 27 & 28 & 33 \\
\maetok, stochastic & 39 & 41 & 43 & 47 \\
\ours, stochastic & 70 & 75 & 79 & 87 \\
\bottomrule
\end{tabular}
\end{table}

\begin{figure}[!htbp]
  \centering
  \includegraphics[width=\linewidth]{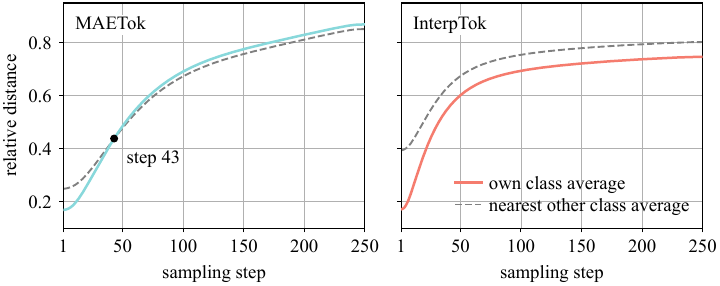}
  \caption{\textbf{The generator's prediction over the sampling steps.} The solid line is the distance from the generator's prediction of the finished code to the average code of its own class, and the dashed line the distance to the nearest average code of another class, at each of 250 steps of the deterministic sampler, averaged over 50{,}048 generations per generator. Both are divided by the mean distance of the finished codes from the average of all validation codes. The dot marks the step from which the prediction of \maetok lies farther from the average code of its own class than from the nearest average code of another class.}
  \label{fig:mech_steps}
\end{figure}

\textbf{Early and late steps.} Furthermore, we want to check the mechanism of our hypothesis, which states that early on, when the noise is strong, the class condition is the stronger cue, and that only in later steps do the patterns in the noise steer the generation more strongly. We therefore expect that in earlier steps, the generator moves the noisy code closer to its class average, and that in later steps it moves it away, so that the code is farther away from its class average and the average of another class becomes relatively closer. To check this, we measure the generator's prediction of the finished code at each of the 250 steps for \ours and for \maetok, and compare its distance to the average code of its own class with its distance to the nearest class average that is not its own, Fig.~\ref{fig:mech_steps}. Table~\ref{tab:mech_traj} gives the values after the first and after the last step. After one step, the prediction of \ours is on average 2.3 times closer to the average code of its own class than to the nearest class average that is not its own. For \maetok, it is 1.5 times closer. After 250 steps, the codes of \ours are only 7\% closer to the average code of their own class than to the nearest class average that is not their own, and the codes of \maetok are on average about as far from the average code of their own class as from the nearest class average that is not their own. Also in absolute terms, the distance of the prediction to the average code of its own class grows with every step the generator takes, as can be seen in Fig.~\ref{fig:mech_steps}. The first step is the part that tests our hypothesis, since there the prediction of both generators lies closer to the average code of its own class than to that of any other class. That the prediction then moves away from the class average is expected, since it has to end at one particular code. We interpret this to mean that early on, when the noise is high, the generator relies more on the class condition and moves toward the average of the codes of that class, and that later it relies more on the patterns of the code, which steer it away from that average. This is our interpretation of the two generators we measured, not a claim about how diffusion models behave in general. Studies of unconditional diffusion models, which get no class label, also find that the class of an image is settled at high noise and its details at lower noise~\citep{biroli24regimes,sclocchi25phase}. If the class average comes first and the details later, this would explain why fewer steps bring the generated codes closer to their class average. With fewer and bigger steps, the generation moves further toward the early prediction, the class average, and fewer steps remain in which the later predictions can steer it away from that average.

\begin{table}[!htbp]
\centering
\begin{minipage}[t]{0.53\textwidth}
  \centering\scriptsize\setlength{\tabcolsep}{3pt}
  \caption{\textbf{The generator's prediction at the first and the last sampling step.} Mean distance from the prediction of the finished code to the reference in each column, in the normalized latent space of the generator. Other average is the nearest average code of another class, and all codes is the average of all validation codes. Deterministic sampler at 250 steps, 50{,}048 starts per generator.}
  \label{tab:mech_traj}
  \begin{tabular}{@{}rrrrrr@{}}
    \toprule
    Step & Noise & Own average & Other average & All codes & Final code \\
    \midrule
    \multicolumn{6}{@{}l}{\textit{the generator on \maetok's codes}} \\
    1 & 1.00 & 5.3 & 7.8 & 15.6 & 26.3 \\
    250 & 0.004 & 27.2 & 26.7 & 31.3 & 0.0 \\
    \midrule
    \multicolumn{6}{@{}l}{\textit{\ours's generator}} \\
    1 & 1.00 & 10.6 & 24.3 & 39.7 & 44.1 \\
    250 & 0.004 & 46.2 & 49.7 & 61.9 & 0.0 \\
    \bottomrule
  \end{tabular}
\end{minipage}\hfill
\begin{minipage}[t]{0.45\textwidth}
  \centering\scriptsize\setlength{\tabcolsep}{3pt}
  \caption{\textbf{Decoded averages of two validation codes.} Equal-weight averages decoded by the named decoder, $n = 50{,}000$ per cell. Each row averages the codes of its own tokenizer, and the indented row averages \maetok's codes and changes only the decoder. Class accuracy is the share of the decoded images that a ResNet-50 trained on ImageNet assigns to the class of the first code.}
  \label{tab:mech_mix}
  \begin{tabular}{@{}lrrrr@{}}
    \toprule
    & \multicolumn{2}{c}{Same class} & \multicolumn{2}{c}{Random classes} \\
    \cmidrule(lr){2-3}\cmidrule(lr){4-5}
    Decoder & FID\,$\downarrow$ & Class acc.\,$\uparrow$ & FID\,$\downarrow$ & Class acc.\,$\uparrow$ \\
    \midrule
    \maetok & 44.6 & 50 & 101.0 & 8 \\
    \quad $+$ our decoder & 7.7 & 81 & 29.5 & 29 \\
    without interp. & 43.3 & 50 & 93.8 & 10 \\
    \ours & \textbf{7.4} & \textbf{85} & \textbf{20.7} & \textbf{36} \\
    \bottomrule
  \end{tabular}
\end{minipage}
\end{table}

\textbf{Nearest codes.} Why should landing closer to the class average help? In our interpretation, with many steps, the generation follows patterns that it thinks it sees in the noise, and these can be patterns of images of any class, so the image may resemble its class less decisively. If a code is sampled around the class average, how much it resembles this class depends on how many of the codes that lie near the class average belong to that class rather than to other classes. We therefore generate about 50 images per class with 250, 32, 16 and 8 steps. For each generated code, we then look in the ImageNet validation set for the 10 closest codes encoded by the tokenizer. Then we see how many of them belong to the same class as the class the generated code used as condition. As can be seen in Table~\ref{tab:mech_gen}, with fewer steps, the generated codes of both \ours and \maetok have more codes of the same class among their nearest codes. Furthermore, \ours consistently has about three times as many codes of the same class among its nearest codes as \maetok. With the stochastic sampler, the numbers are almost the same. So when the generated codes of \ours land closer to their class average, they land among codes of their own class far more often than those of \maetok. This indicates that the codes of \ours that land around the class average resemble mixtures of codes of their class more, while those of \maetok resemble mixtures that include more codes of other classes.

\begin{table}[!htbp]
\centering\scriptsize\setlength{\tabcolsep}{4pt}
\caption{\textbf{Generated codes at four numbers of sampling steps.} 50{,}048 generated codes per generator in blocks a to e and 50{,}000 in blocks f to i and in the stochastic rows of block b, deterministic sampler unless the block or row says otherwise. Block d is the distance of the generated code at 8 steps from the generated code at 250 steps from the same starting noise, in percent of the length of the code.}
\label{tab:mech_gen}
\begin{tabular}{@{}lrrrr@{}}
\toprule
Sampling steps & 250 & 32 & 16 & 8 \\
\midrule
\multicolumn{5}{@{}l}{\textit{a. within-class spread, relative to 250 steps}} \\
\maetok & 1 & 0.96 & 0.92 & 0.84 \\
\ours & 1 & 0.95 & 0.90 & 0.79 \\
without interp. & 1 & 0.95 & 0.89 & 0.78 \\
without reg. & 1 & 0.96 & 0.91 & 0.81 \\
\midrule
\multicolumn{5}{@{}l}{\textit{b. share of the ten nearest validation codes that carry the code's label, percent}} \\
\maetok & 18 & 19 & 20 & 21 \\
\ours & 55 & 58 & 61 & 66 \\
without interp. & 59 & 62 & 65 & 72 \\
without reg. & 10 & 10 & 10 & 10 \\
\maetok, stochastic & 19 & 19 & 20 & 21 \\
\ours, stochastic & 55 & 58 & 60 & 66 \\
\midrule
\multicolumn{5}{@{}l}{\textit{c. distance to the nearest validation code, relative to the spacing of validation codes}} \\
\maetok & 1.00 & 0.98 & 0.96 & 0.92 \\
\ours & 0.99 & 0.97 & 0.95 & 0.90 \\
without interp. & 1.01 & 0.99 & 0.96 & 0.92 \\
without reg. & 1.00 & 0.98 & 0.95 & 0.89 \\
\midrule
\multicolumn{5}{@{}l}{\textit{d. distance of the code at 8 steps from the code at 250 steps, percent of the length of the code}} \\
\maetok & \multicolumn{4}{r@{}}{13} \\
\ours & \multicolumn{4}{r@{}}{26} \\
without interp. & \multicolumn{4}{r@{}}{23} \\
without reg. & \multicolumn{4}{r@{}}{10} \\
\bottomrule
\end{tabular}
\par\medskip
\begin{minipage}[t]{0.56\textwidth}\centering\scriptsize\setlength{\tabcolsep}{4pt}
\begin{tabular}{@{}lrr@{}}
\toprule
Sampling steps & 250 & 32 \\
\midrule
\multicolumn{3}{@{}p{7.1cm}@{}}{\textit{e. within-class spread with the stochastic sampler, relative to 250 steps of the deterministic sampler}} \\
\maetok & 0.99 & 0.97 \\
\ours & 0.99 & 0.95 \\
\midrule
\multicolumn{3}{@{}p{7.1cm}@{}}{\textit{f. distance to the nearest validation code with the stochastic sampler, relative to the spacing of validation codes}} \\
\maetok & 0.99 & 0.98 \\
\ours & 0.98 & 0.97 \\
\midrule
\multicolumn{3}{@{}p{7.1cm}@{}}{\textit{g. distance to the nearest segment between two of the ten nearest validation codes, relative to the distance to the nearest one}} \\
\maetok & 0.84 & 0.84 \\
\ours & 0.83 & 0.83 \\
\bottomrule
\end{tabular}
\end{minipage}\hfill
\begin{minipage}[t]{0.42\textwidth}\centering\scriptsize\setlength{\tabcolsep}{4pt}
\begin{tabular}{@{}lrr@{}}
\toprule
Codes that are not generated & \maetok & \ours \\
\midrule
\multicolumn{3}{@{}p{5.3cm}@{}}{\textit{h. as in block g, where moved at random means a validation code moved in a random direction by a generated code's distance to its nearest validation code}} \\
a validation code & 0.85 & 0.83 \\
moved at random & 0.93 & 0.92 \\
\midrule
\multicolumn{3}{@{}p{5.3cm}@{}}{\textit{i. as in block c, absolute}} \\
a validation code & 31 & 50 \\
moved at random & 44 & 71 \\
\bottomrule
\end{tabular}
\end{minipage}
\end{table}

\textbf{The encoder.} We attribute the fact that the generated codes of \ours have about three times as many codes of their class around them to the encoder, since the encoder decides where the codes of the images lie. For the codes of the ImageNet validation images, we check how often a code is closer to the average of the codes of its own class than to the averages of all other classes, Table~\ref{tab:mech_latent}. With \ours's encoder, this is the case for 73\% of the codes, with \maetok's for 55\%. Furthermore, of the 100 codes closest to a class average, on average 37 are of that class with \ours's encoder and 17 with \maetok's, so twice as many. The variant trained without interpolation is about as good as \ours here, with 73\% of its codes closest to their own class average and on average 39 codes of the class among the 100 codes closest to a class average. The variant trained without the regularization losses is the worst, with only 43\% of its codes closest to their own class average and only 5 codes of the class among the 100 closest, so this property comes mostly from the regularization losses. \ours's encoder therefore makes it more likely that a code near its class average is really a code of that class.

\begin{table}[!htbp]
\centering\scriptsize\setlength{\tabcolsep}{4pt}
\caption{\textbf{The latent space of each encoder.} The codes of the 50{,}000 ImageNet validation images, in the normalized latent space of the generator unless a block says otherwise. The linear probe follows the MAE procedure of Table~\ref{tab:protocol}. Without interp.\ and without reg.\ are the rows w/ reg.\ and w/ interpolation of Table~\ref{tab:abl}. $^{\dagger}$Quoted from \citet{zheng26rae} under its own procedure.}
\label{tab:mech_latent}
\begin{tabular}{@{}lrrrr@{}}
\toprule
& \maetok & \ours & \begin{tabular}[b]{@{}r@{}}without\\interp.\end{tabular} & \begin{tabular}[b]{@{}r@{}}without\\reg.\end{tabular} \\
\midrule
Linear-probe accuracy, percent & 73.0 & 77.9 & 77.0 & 63.5 \\
Nearest neighbor of the same class, percent & 30 & 61 & 65 & 17 \\
Nearest-neighbor distance, relative & 1 & 1.6 & 1.6 & 0.9 \\
Own-class codes among the 100 nearest to a class average, of at most 50 & 17 & 37 & 39 & 5 \\
Nearest class average is its own, percent & 55 & 73 & 73 & 43 \\
\midrule
\multicolumn{5}{@{}l}{\textit{in the unnormalized latent space of the decoder}} \\
Per-dimension variation, relative & 1 & 2.2 & 2.1 & 0.8 \\
Effective dimensions, the participation ratio of the covariance spectrum, relative & 1 & 2.1 & 3.1 & 1.2 \\
\bottomrule
\end{tabular}
\par\medskip
\begin{minipage}[t]{0.52\textwidth}\centering\scriptsize\setlength{\tabcolsep}{4pt}
\begin{tabular}{@{}lrr@{}}
\toprule
& \maetok & \ours \\
\midrule
Between-class to within-class spread, relative & 1 & 2 \\
\bottomrule
\end{tabular}
\end{minipage}\hfill
\begin{minipage}[t]{0.44\textwidth}\centering\scriptsize\setlength{\tabcolsep}{4pt}
\begin{tabular}{@{}lr@{}}
\toprule
Two further encoders & Linear probe \\
\midrule
SD-VAE & 4.9 \\
RAE$^{\dagger}$ & 84.5 \\
\bottomrule
\end{tabular}
\end{minipage}
\end{table}

\textbf{The decoder.} Furthermore, the class average is the average of many codes, so a sampled code near it resembles a mixture of many codes more than the code of a single image. \ours's decoder was trained to decode mixtures of codes, while \maetok's decoder was never trained to handle mixtures of codes. To check whether \ours's decoder therefore decodes such codes better, we average the codes of two, three, five and ten consecutive validation images of one class and decode the average with both decoders, Table~\ref{tab:kway}. We measure the Inception Score of the decoded images~\citep{salimans16is}, a standard measure that is higher the more confident an image classifier is about which class each image shows, and lower for blends of several images. As a reference, decoding the codes of single real images gives an Inception Score of about 140 with both decoders. For averages of two codes, \ours's decoder still reaches 139, \maetok's decoder only 48. For averages of three and five codes, \ours's decoder reaches 131 and 92, \maetok's decoder only 25 and 11. For averages of two such codes, the FID of the decoded images is 3.79 with \ours's decoder and 35.12 with \maetok's decoder, Table~\ref{tab:kway}. Table~\ref{tab:mech_mix} repeats this with other pairs of codes of one class and adds two more decoders. There, the FID is 7.4 with \ours's decoder and 44.6 with \maetok's decoder, six times lower. Our decoder on \maetok's codes, trained with \maetok's encoder frozen, reaches 7.7, so most of the gain comes from the decoder. The variant trained without interpolation, whose decoder was not trained on mixtures of codes, is about as bad as \maetok, with 43.3. So \ours's decoder turns codes that land near the class average into more natural images than \maetok's decoder, which turns them into blends more often.

\textbf{Generated images.} Finally, we check whether, with fewer steps, more of the generated images show the class they were generated for, since this is what our explanation predicts for the images and what FID rewards. The measure we use for generation quality, FID, is how well the generated images, as seen by a vision model, resemble the distribution of the dataset~\citep{heusel17fid}. This is the standard metric for generation quality, and it is lower the more natural the images look and the more closely they follow the class distribution. For about 50 generated images per class, a standard ImageNet classifier, a ResNet-50, says which class each image shows, and we count how often it says the class the image was generated for, Table~\ref{tab:mech_class}. With the deterministic sampler, from 250 to 8 steps, this rises for \ours from 61\% to 83\% and for \maetok's codes with our decoder from 62\% to 71\%, while \maetok stays at about 60\%. With the stochastic sampler, the share rises for \ours from 68\% to 71\% at 32 steps and falls slightly for \maetok, from 64\% to 63\%. So with fewer steps, more of the images of \ours show the class they were made for.

\begin{table}[!htbp]
\centering\scriptsize\setlength{\tabcolsep}{4pt}
\caption{\textbf{Class accuracy and similarity of the decoded images}, 50{,}000 decoded images per cell, deterministic sampler unless a block says otherwise. Class acc.\ is the share of the decoded images whose top-1 class under a ResNet-50 trained on ImageNet is the conditioned label, and sim.\ is the mean cosine similarity of the features of those images within a class.}
\label{tab:mech_class}
\begin{tabular}{@{}lrrrrrrrr@{}}
\toprule
& \multicolumn{4}{c}{Class acc., percent} & \multicolumn{4}{c}{Sim.} \\
\cmidrule(lr){2-5}\cmidrule(lr){6-9}
Sampling steps & 250 & 32 & 16 & 8 & 250 & 32 & 16 & 8 \\
\midrule
\maetok & 60 & 60 & 61 & 58 & 0.42 & 0.43 & 0.43 & 0.45 \\
\quad $+$ our decoder & 62 & 64 & 66 & 71 & 0.42 & 0.43 & 0.44 & 0.47 \\
\ours & 61 & 67 & 73 & 83 & 0.43 & 0.45 & 0.48 & 0.56 \\
\bottomrule
\end{tabular}
\par\medskip
\begin{minipage}[t]{0.50\textwidth}\centering\scriptsize\setlength{\tabcolsep}{3pt}
\begin{tabular}{@{}lrrrrrrrr@{}}
\toprule
& \multicolumn{4}{c}{Class acc., percent} & \multicolumn{4}{c}{Sim.} \\
\cmidrule(lr){2-5}\cmidrule(lr){6-9}
Sampling steps & 250 & 32 & 16 & 8 & 250 & 32 & 16 & 8 \\
\midrule
without interp. & 49 & 53 & 56 & 57 & 0.41 & 0.43 & 0.45 & 0.48 \\
without reg. & 50 & 52 & 55 & 61 & 0.40 & 0.41 & 0.42 & 0.45 \\
\bottomrule
\end{tabular}
\end{minipage}\hfill
\begin{minipage}[t]{0.48\textwidth}\centering\scriptsize\setlength{\tabcolsep}{4pt}
\begin{tabular}{@{}lrrrr@{}}
\toprule
& \multicolumn{2}{c}{Class acc., percent} & \multicolumn{2}{c}{Sim.} \\
\cmidrule(lr){2-3}\cmidrule(lr){4-5}
Sampling steps & 250 & 32 & 250 & 32 \\
\midrule
\multicolumn{5}{@{}p{6.2cm}@{}}{\textit{with the stochastic sampler, new draws whose gFID matches Table~\ref{tab:nfe} within 0.2}} \\
\maetok & 64 & 63 & 0.43 & 0.43 \\
\quad $+$ our decoder & 67 & 66 & 0.43 & 0.44 \\
\ours & 68 & 71 & 0.44 & 0.47 \\
\bottomrule
\end{tabular}
\end{minipage}
\end{table}

\textbf{Summary.} We read these experiments as support for our hypothesis on our generators, without claiming that they prove it or that it describes generative models in general. We show that with fewer sampling steps, where each step is bigger, the generated codes lie closer to the average code of the class used as condition for generation. We furthermore show that within 250 steps, at the start the prediction is closer to the class average and later moves away from it. Furthermore, we show that codes that lie around the class average are far more likely to be of the same class for \ours than for \maetok. We therefore hypothesize that the following causes the observation. With fewer steps, the generated codes are closer to the class average. Codes around the class average more often resemble that class for \ours. Furthermore, \ours was trained to decode codes that are mixtures of codes, and it decodes them into more natural images. Together, with fewer steps, the generated codes of \ours move toward their own class average more than toward the averages of other classes, and \ours's decoder turns them into more natural images that look more like their class than with \maetok. Since FID compares the distribution of the generations with that of the dataset as viewed by a vision model, it rewards generations that adhere more closely to the class condition and have higher visual fidelity. This mechanism works less well for the two variants. For the variant trained without the regularization losses, codes near the class average were even less often of that class than with \maetok, Table~\ref{tab:mech_latent}. The variant trained without interpolation had classes about as distinguishable as \ours, Table~\ref{tab:mech_latent}, but its decoder turns mixtures of codes into images about as unnatural as \maetok's decoder, Table~\ref{tab:mech_mix}. Our decoder on \maetok's codes decodes mixtures of codes about as well as \ours's decoder, but \maetok's codes around the class average are less often of that class. Without the added noise, these models still improve with fewer steps, but less than \ours, and with the stochastic sampler, they get worse, Table~\ref{tab:mech_fid}.

%% file: appendix/J_controls.tex
\subsection{A second training seed}
\label{app:results:seeds}

Each SiT-B/1D trains for 400{,}000 steps at batch 256 with seed 0, and a second generator of each tokenizer, trained with seed 1, reproduces the gap, Table~\ref{tab:seeds}.

\begin{table}[!htb]
  \centering\footnotesize\setlength{\tabcolsep}{6pt}
  \caption{\textbf{A second training seed.} SiT-B/1D after 400{,}000 steps, 250 sampling steps without guidance, $n = 50{,}000$ per cell, one sampling run per generator. Seed 0 is the generator of Table~\ref{tab:nfe}, whose draw there gives 11.61 and 8.48.}
  \label{tab:seeds}
  \begin{tabular}{@{}lrr@{}}
    \toprule
    \gfid\,$\downarrow$ & seed 0 & seed 1 \\
    \midrule
    \maetok & 11.51 & 11.38 \\
    \ours & \textbf{8.51} & \textbf{8.28} \\
    \bottomrule
  \end{tabular}
\end{table}

\subsection{Further draws of the released SiT-XL}
\label{app:results:swap_xl}

Two further sets of samples of the released SiT-XL, drawn with other seeds, give 2.44 and 2.49 with \maetok's decoder and 2.24 with ours on both, Table~\ref{tab:swap_xl}.
FD-DINOv2 is FID computed in DINOv2 features instead of Inception features~\citep{oquab24dinov2,stein23exposing}.
A decoder trained further with \maetok's own losses decodes them with a higher gFID than \maetok's decoder at every number of sampling steps, while ours lowers it at every one.
At 13{,}000 fine-tuning steps, our decoder without the six appended blocks matches ours with them, removing the feature mixing ratio loss raises the gFID to 2.50, above \maetok's 2.44, and \maetok's losses with noise augmentation reach 2.70.

\begin{table}[!htb]
  \centering\footnotesize\setlength{\tabcolsep}{2.5pt}
  \caption{\textbf{The released SiT-XL of \citet{chen25maetok} decoded by each decoder.} \gfid at 250, 64, 32 and 16 sampling steps, FD-DINOv2 and CMMD~\citep{jayasumana24cmmd} at 250 steps, and the \gfid of a second draw at 250 steps. Every decoder keeps \maetok's frozen encoder. The draw of Table~\ref{tab:nfe} gives 2.46 for \maetok's decoder and 2.26 for ours. A dash marks a value that is not measured. Bold is the best value in its column among \maetok's decoder and the 41{,}000-step decoders. $n = 50{,}000$ per cell.}
  \label{tab:swap_xl}
  \begin{tabular}{@{}lrrrrrrr@{}}
    \toprule
    Decoder & 250 & 64 & 32 & 16 & FD-DINOv2\,$\downarrow$ & CMMD\,$\downarrow$ & 2nd draw \\
    \midrule
    \maetok & 2.44 & 2.76 & 3.44 & 7.49 & 125.6 & 0.313 & 2.49 \\
    \midrule
    \multicolumn{8}{@{}l}{\emph{41{,}000 fine-tuning steps}} \\
    \quad \maetok's losses & 2.61 & 2.88 & 3.55 & 7.81 & 108.2 & 0.226 & 2.63 \\
    \quad \ours & \textbf{2.24} & \textbf{2.39} & \textbf{2.79} & \textbf{5.80} & \textbf{88.1} & \textbf{0.171} & \textbf{2.24} \\
    \midrule
    \multicolumn{8}{@{}l}{\emph{13{,}000 fine-tuning steps}} \\
    \quad \maetok's losses w/ noise augmentation & 2.70 & -- & -- & -- & 124.4 & 0.258 & -- \\
    \quad \ours & 2.30 & 2.45 & 2.87 & 5.71 & 89.2 & 0.177 & -- \\
    \qquad w/o six appended blocks & 2.30 & 2.44 & 2.86 & 5.68 & 89.3 & 0.177 & -- \\
    \qquad w/o repr.\ Fr\'echet loss & 2.29 & -- & -- & -- & 93.6 & 0.189 & -- \\
    \qquad w/o Inception & 2.29 & -- & -- & -- & -- & -- & -- \\
    \qquad w/o pixel mixing ratio loss & 2.32 & -- & -- & -- & -- & -- & -- \\
    \qquad w/o feature mixing ratio loss & 2.50 & -- & -- & -- & -- & -- & -- \\
    \qquad w/o source conditioning & 2.31 & -- & -- & -- & 89.2 & 0.177 & -- \\
    \bottomrule
  \end{tabular}
\end{table}

\subsectionnobarrier{Uncurated samples}
\label{app:results:uncurated}

\begin{figure}[H]
  \centering
  \includegraphics[width=0.49\textwidth]{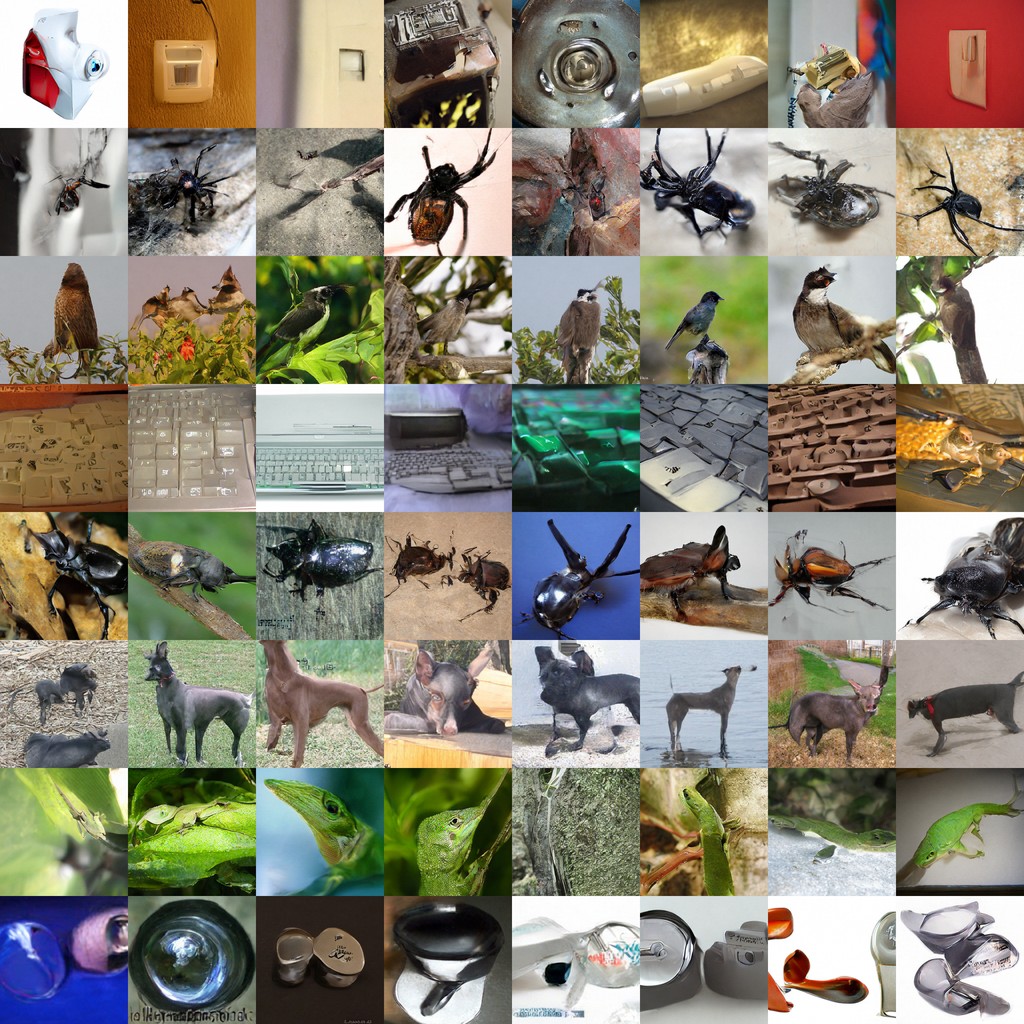}\hfill
  \includegraphics[width=0.49\textwidth]{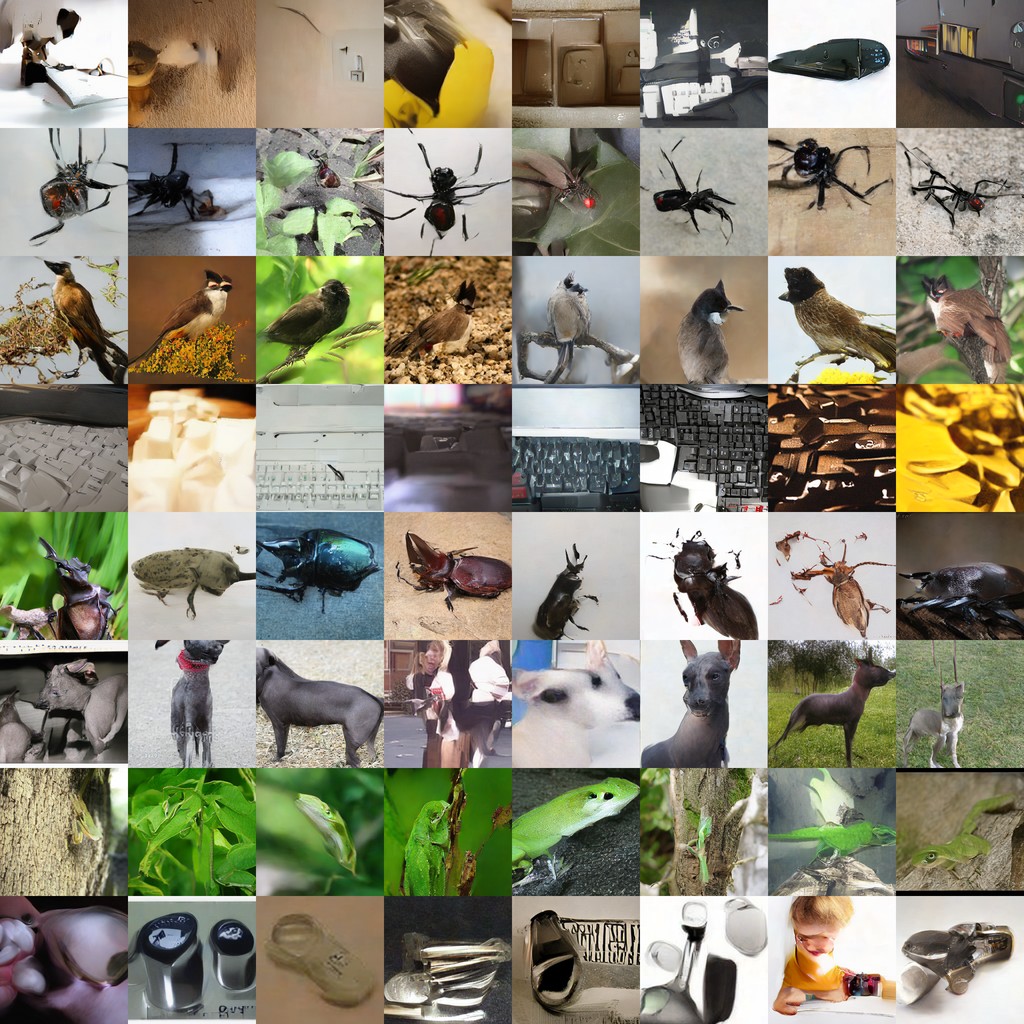}
  \caption{\textbf{Uncurated samples.} SiT-B on \maetok's codes on the left and on \ours's codes on the right, 250 steps of the stochastic sampler without guidance. One class per row, switch, black widow, bulbul, computer keyboard, rhinoceros beetle, Mexican hairless, American chameleon and loupe, with the first eight samples of each.}
  \label{fig:uncurated}
\end{figure}

Fig.~\ref{fig:uncurated} shows uncurated samples of both SiT-B generators.
The classes are drawn at random with seed 0, and every row shows the first eight samples of its class.

%% file: appendix/H_smallcode.tex
\section{Interpolation under a Smaller Code}
\label{app:smallcode}

We ran an earlier experiment with a much smaller code, trained on ImageNet and the image-editing pairs of ImgEdit~\citep{ye25imgedit}, and record what it produced beside the main result.

\begin{figure}[!htb]
  \centering
  \includegraphics[width=0.90\linewidth]{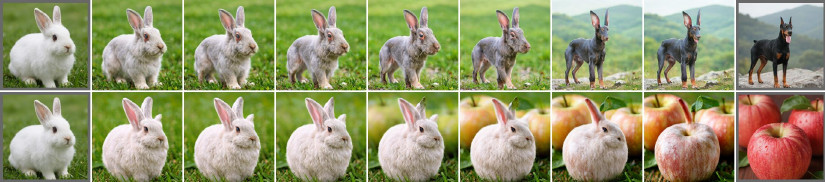}
  \caption{\textbf{Decoded paths under a smaller code.} In each row, the two dark-outlined ends decode the codes of two images generated with Z-Image-Turbo~\citep{zimage25} and the seven tiles between them mixtures of those two codes.}
  \label{fig:smallcode_rows}
\end{figure}

\begin{figure}[!tp]
  \centering
  \includegraphics[width=0.70\linewidth]{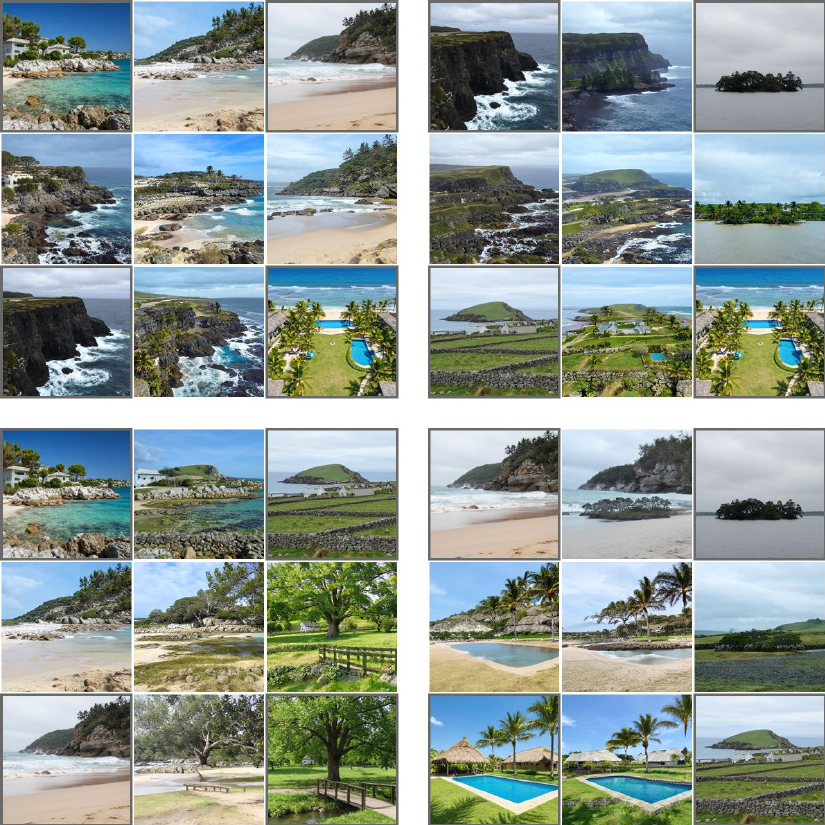}
  \caption{\textbf{Decoded mixtures of four codes under a smaller code.} Four unrelated squares, one per quadrant, each tile decoded from a mixture of its four corner codes with weights bilinear in the position. Only the dark-outlined corners hold images generated with Z-Image-Turbo~\citep{zimage25}.}
  \label{fig:smallcode_box}
\end{figure}

The tokenizer is a frozen REPA-E end-to-end autoencoder~\citep{leng25repae} read by a transformer encoder at SiT-B scale and written back by a flow decoder at SiT-L scale~\citep{ma24sit}.
Only those two are trained, and they carry the code through thirty-two register tokens of thirty-two dimensions each, normalized together onto a sphere.
Some other implementation details differ from \ours as well.

The smaller code is more semantic, and its mixtures change the content of the scene more and superimpose the two images less, a mixture of a beach and a cliff being one coastline.
Fig.~\ref{fig:smallcode_rows} shows the decoded images at positions along the line between two image codes, and Fig.~\ref{fig:smallcode_box} decodes mixtures of four codes placed at the corners of a square.

\input{tables/H1_smallcode.tex}
Reconstruction is the cost, and Table~\ref{tab:smallcode} sets it beside \maetok and \ours.

One variant sampled random codes alongside the interpolated ones, since a smaller and more semantic code should make that easier, and it collapsed, in the way a generative adversarial network does, onto a single incoherent image, decoding a meaningful image from an arbitrary code being a more ambiguous task than decoding one between two image codes.

A smaller bottleneck can make the codes more semantic at the cost of reconstruction, and for semantic interpolation it can be the better choice, Figs.~\ref{fig:smallcode_rows} and~\ref{fig:smallcode_box}.

%% file: tables/H1_smallcode.tex
\Needspace{9\baselineskip}
\begin{wraptable}[8]{r}{0.48\textwidth}
\vspace{-\intextsep}
\caption{\textbf{Reconstruction under a smaller code.} ImageNet, 50{,}000 images, SSIM in \maetok's convention.}
\label{tab:smallcode}
\centering\footnotesize\setlength{\tabcolsep}{4pt}
\begin{tabular}{@{}lrrr@{}}
\toprule
Tokenizer & PSNR\,$\uparrow$ & SSIM\,$\uparrow$ & \lpips\,$\downarrow$ \\
\midrule
\ours & 24.02 & 0.775 & 0.0819 \\
\maetok & 23.55 & 0.760 & 0.0917 \\
Smaller code & 19.31 & 0.615 & 0.1735 \\
\bottomrule
\end{tabular}
\end{wraptable}

%% file: appendix/A_method_details.tex
\sectionnobarrier{Experimental Setup}
\label{app:method}

\subsection{Training Details of the Tokenizer}
\label{app:method:hparams}

Table~\ref{tab:config} gives the training configuration.
\maetok masks 40 to 60\% of the image patches and trains three auxiliary decoders that regress histogram of oriented gradients (HOG), DINOv2-L and SigLIP-SO400M features of the masked patches from the code.
A second stage freezes the encoder and fine-tunes the pixel decoder under reconstruction, perceptual and adversarial losses~\citep{chen25maetok}.
Our fine-tune uses neither masking nor auxiliary decoders. It trains the encoder as well as the decoder, since reconstruction alone does not fix the shape of the latent space. Reshaping the latent space with an invertible map and undoing the map in the decoder leaves every reconstruction unchanged, but it changes the straight line between two codes unless the map is affine. Two tokenizers that reconstruct identically can therefore decode the line between two images differently, and the encoder decides which codes lie on this line.
The CLS feature of the frozen vision model and a list of the 64 nearest neighbors are precomputed for every training image.
The variants of Sec.~\ref{sec:ablations} are compared at a matched number of steps, 41{,}000 for those that train encoder and decoder and 13{,}000 for those that train only the decoder, Table~\ref{tab:abl}.

\input{tables/A1_config.tex}

\subsectionnobarrier{Training Objectives}
\label{app:method:losses}

A micro-batch is the part of a batch that one forward pass processes.
It contains $n$ single images and $m$ mixtures, as many as Table~\ref{tab:config} states, and every pair of its single images gives one mixture.
Every fine-tuning step therefore trains reconstruction at the codes of training images and the interpolation objective between them, with three and a half times as many mixtures as single images.
Each loss below is listed with its formula, its weight at the start of training and the images it acts on, single images or mixtures.
For an image $X$, $D(E(X))$ is its reconstruction, $ab_\alpha$ and $AB_\alpha$ are as in Sec.~\ref{sec:method}, and $f_A$, $f_B$ and $f_{AB}$ are the unit CLS features of $A$, $B$ and $AB_\alpha$ in the frozen vision model, DINOv3-B/16.

\begin{enumerate}[leftmargin=1.5em]
\item \textbf{Reconstruction} $\mathcal{L}_{\mathrm{rec}} = \tfrac1n\sum_X \big(\|D(E(X)) - X\|_1 + \mathrm{LPIPS}_{\mathrm{VGG}}(D(E(X)), X)\big)$~\citep{zhang18lpips} acts on single images at weights 1.0 and 1.0 and is the numerator of the adaptive adversarial weight.

\item \textbf{Generator adversarial} $\mathcal{L}_G = \mathcal{L}_G^{\mathrm{int}} + 1.0\cdot\mathcal{L}_G^{\mathrm{rec}}$ acts on both at weight $w = w_{\mathrm{ad}}\cdot 0.4\cdot g_{\mathrm{gate}}$, with $\mathcal{L}_G^{\mathrm{int}}$ of Eq.~\ref{eq:slot}, $\mathcal{L}_G^{\mathrm{rec}} = -\mathbb{E}[\text{logit}]$, $w_{\mathrm{ad}} = \|\nabla \mathcal{L}_{\mathrm{rec}}\| / (\|\nabla \mathcal{L}_G\| + 10^{-4})$ taken at the decoder's final linear layer and capped at 1.0, and $g_{\mathrm{gate}} = 0$ for the first 1{,}500 steps.
The alignment losses do not enter this ratio. The discriminators train from step 0, whereas \maetok starts its discriminator at step 30{,}000.

\item \textbf{Discriminator} $\mathcal{L}_D = 0.4\,[\mathcal{L}_D^{\mathrm{int}} + 1.0\cdot\mathcal{L}_D^{\mathrm{rec}} + 10^{-3}\,\mathcal{L}_{\mathrm{LeCAM}}] + 4.0\,\mathcal{L}_{\mathrm{CR}}$ uses $\mathcal{L}_D^{\mathrm{int}}$ of Eq.~\ref{eq:slot} and $\mathcal{L}_D^{\mathrm{rec}} = \tfrac12\,\mathbb{E}[(1-\text{real})_+ + (1+\text{fake})_+]$ per logit on real images against their reconstructions.
LeCAM~\citep{tseng21lecam} uses EMA decay 0.999, consistency regularization~\citep{zhang20cr} compares a DiffAug view with a cutout view, and DiffAug~\citep{zhao20diffaug} is applied to the inputs of both networks at probability 0.5.

\item \textbf{Pixel mixing ratio loss} $\mathcal{L}_{\mathrm{ratio}}$ of Eq.~\ref{eq:ratio} acts on mixtures at weight 2.0, averaged over the mixtures, with the reconstructions $D(a)$ and $D(b)$ of the same forward pass treated as fixed targets without gradient.
We do not pull the decoded mixture toward the pixel blend $(1-\alpha)D(a) + \alpha D(b)$ instead, since the only image that minimizes that loss is the blend itself, the crossfade the interpolation objective is meant to avoid. Eq.~\ref{eq:ratio} constrains only the ratio of the two distances, which many images satisfy, and the interpolation discriminator selects a natural one among them.

\item \textbf{Feature mixing ratio loss} $\mathcal{L}_{\mathrm{ratio}}^{\mathrm{feat}}$ acts at weight 0.5 on at most 8 of the 28 mixtures of a micro-batch, whose decoded images pass through the frozen vision model with gradient.
With $\theta(u,v)$ the angle between $u$ and $v$,
\begin{equation}
\label{eq:arccos}
\mathcal{L}_{\mathrm{ratio}}^{\mathrm{feat}} = \mathbb{E}_\alpha\Big[\big(\theta(f_{AB},f_A) - \alpha\,\theta(f_A,f_B)\big)^2 + \big(\theta(f_{AB},f_B) - (1-\alpha)\,\theta(f_A,f_B)\big)^2\Big],
\end{equation}
whose zero requires $\theta(f_{AB},f_A) + \theta(f_{AB},f_B) = \theta(f_A,f_B)$ and places $f_{AB}$ on the shorter great-circle arc from $f_A$ to $f_B$.

\item \textbf{Representation Fr\'echet loss} $\mathcal{L}_{\mathrm{fd}} = \sum_e \mathrm{FD}_e / (\mathrm{sg}(\mathrm{FD}_e) + 0.01)$, with $\mathrm{sg}$ the stop-gradient, acts at weight 0.3 on an 8-image subsample of the 36 decoded images of a micro-batch, single images and mixtures, over the frozen SigLIP2-L/16 pooler, the Inception-v3 pool features and the MAE ViT-L/16 CLS feature.
Inception-v3 is also the feature space of FID.
The loss follows \citet{yang26fdloss} at EMA decay 0.999 with an eigenvalue matrix root and a stop-gradient normalizer, on reference moments of the ImageNet training images, after the moments have accumulated over 500 optimizer steps.

\item \textbf{Distance-matrix alignment} $\mathcal{L}_{\mathrm{axis}} = \mathrm{MSE}(\hat z \hat z^\top, \hat t\, \hat t^\top)$ acts on single images at weight 0.5, where $\hat z$ holds the $n\times 4096$ row-normalized flattened codes and $\hat t$ the $n\times 768$ row-normalized CLS features of the frozen vision model.
It covers the full $8\times 8$ matrix, diagonal included, and adapts the distance-matrix loss of VA-VAE~\citep{yao25vavae}, which compares the positions within one image, to pairs of images.

\item \textbf{Latent-CLS alignment} $\mathcal{L}_{\mathrm{pred}} = \mathbb{E}[1 - \cos(h(z), t)]$ acts on single images at weight 0.25, with $h$ a trainable MLP $4096 \to 1024 \to 768$ with SiLU on the flattened code $z$ and $t$ the CLS feature of the frozen vision model, with the MLP projection and cosine objective of REPA~\citep{yu25repa} applied to the code.

\item \textbf{Decoder-patch alignment} $\mathcal{L}_{\mathrm{patch}} = \mathbb{E}_{\mathrm{image},\,\mathrm{pos}}[1 - \cos(h'(u), t')]$ acts on single images at weight 0.1, with $u$ the output of decoder block~6 of~18 at each of the 256 image-token positions, $h'$ a per-position MLP $768 \to 1024 \to 768$ and $t'$ the 256 patch tokens of the frozen vision model without gradient, following the semantic regularization of GigaTok~\citep{xiong25gigatok}.
\end{enumerate}

The full objective of \ours is
\begin{equation}
\label{eq:total}
\mathcal{L} = \underbrace{\mathcal{L}_{\mathrm{rec}} + w\mathcal{L}_G^{\mathrm{rec}} + \lambda_1\mathcal{L}_{\mathrm{axis}} + \lambda_2\mathcal{L}_{\mathrm{pred}} + \lambda_3\mathcal{L}_{\mathrm{patch}}}_{\text{single images}}
+ \underbrace{w\mathcal{L}_G^{\mathrm{int}} + \lambda_4\mathcal{L}_{\mathrm{ratio}} + \lambda_5\mathcal{L}_{\mathrm{ratio}}^{\mathrm{feat}}}_{\text{mixtures}}
+ \underbrace{\lambda_6\mathcal{L}_{\mathrm{fd}}}_{\text{both}},
\end{equation}
where $w$ is the gated adaptive adversarial weight of item 2, after \citet{esser21taming}, and every $\lambda$ is the weight given above.
In the terms of Eq.~\ref{eq:objective}, $\mathcal{L}_{\mathrm{rec}}$ also holds the adversarial term on single images, $\mathcal{L}_{\mathrm{adv}}$ is the adversarial term on mixtures, $\mathcal{L}_{\mathrm{ratio}}$ holds both mixing ratio losses, and $\mathcal{L}_{\mathrm{reg}}$ collects the three alignment losses and the representation Fr\'echet loss.

\textbf{Variants of Table~\ref{tab:abl}.}
The variant trained without the regularization losses starts from the released weights and keeps the images, the pair schedule, the optimizer, the batch and the schedule of the full recipe.
It trains encoder and decoder, without the appended blocks, under the reconstruction objective, the interpolation discriminator and the pixel mixing ratio loss.
The variant trained without interpolation trains encoder and decoder with the reconstruction objective, the three alignment losses and the representation Fr\'echet loss, and it uses the appended decoder blocks.

\subsection{Discriminator Architecture}
\label{app:method:discriminator}

\textbf{Backbone and heads.}
Both discriminators read the features of five layers of the frozen DINO ViT-S/16 backbone~\citep{caron21dino} of \maetok's discriminator, before the first block and after blocks 2, 5, 8 and 11, with the CLS token added to every patch token.
Each layer carries a spectral-normalized~\citep{miyato18sn} head with one logit per token, giving $5\times 196 = 980$ hinge logits per image.

\textbf{Conditioning and passes.}
Each pass of the interpolation discriminator pairs the decoded mixture with one source and scores one of the two given the other.
Before each of its heads, the tokens of the scored image query the keys and values of the other image at the same layer, and a $k{=}1$ convolution fuses $[x;\ \mathrm{att};\ x - \mathrm{att}]$, which makes their difference an explicit input.
The condition thus enters through the features, as in a projection discriminator~\citep{miyato18projection}.
With $s(X\mid Y)$ the per-token logit of image $X$ over the keys of image $Y$, the relativistic comparison of \citet{jolicoeur19relativistic} uses
\begin{equation}
\label{eq:slot}
\begin{gathered}
s_{\mathrm{real}} = \tfrac12\big[s(A \mid AB_\alpha) + s(B \mid AB_\alpha)\big],\qquad
s_{\mathrm{fake}} = \tfrac12\big[s(AB_\alpha \mid A) + s(AB_\alpha \mid B)\big],\\
\mathcal{L}_D^{\mathrm{int}} = \tfrac12\,\mathbb{E}\big[(1 - s_{\mathrm{real}})_+ + (1 + s_{\mathrm{fake}})_+\big],\qquad
\mathcal{L}_G^{\mathrm{int}} = \mathbb{E}[s_{\mathrm{real}}] - \mathbb{E}[s_{\mathrm{fake}}].
\end{gathered}
\end{equation}
Each mixture is thus scored four times, and the two real and the two fake passes are averaged before the hinge.
The reconstruction discriminator is an unconditional head that compares real images with their reconstructions, since a head conditioned on the source would classify any departure from it as fake, and conditioning is known to cause mode collapse in generative adversarial networks~\citep{goodfellow14gan} trained on limited data~\citep{shahbazi22collapse}.
It is separate from the interpolation discriminator, since mixtures are easier to detect as fake than reconstructions and could dominate a shared head.
Table~\ref{tab:config:critic} gives the parameter counts and the schedule.

\subsection{Decoder Architecture}
\label{app:method:growth}

\maetok's decoder has 12 ViT blocks of width 768 over 256 image tokens.
Each of the six appended blocks computes $x + g\,(\mathrm{block}(x) - x)$ with a scalar gate $g$ initialized at zero, and its rotary frequencies are copied from the pretrained decoder.
The appended blocks add 43.1M parameters to the 173.9M of the released checkpoint.

\subsection{Pair Sampling}
\label{app:method:schedule}

In each epoch, every training image is paired with its most similar unpaired image among its $k$ nearest neighbors, measured by the cosine similarity of the CLS features of the frozen vision model. We pair similar images because, by the analysis of \citet{xu26ifid}, at low noise the samples of a diffusion model gather near the straight line between two neighboring training codes. \citet{yue26pae} train the neighborhood of one code, while we train the whole line between two codes.
The number $k$ grows from 8 to 128 over the epochs, Table~\ref{tab:config}, so the pairs become less similar as training proceeds.
Only the 64 nearest neighbors are precomputed, and for $k > 64$ an image takes such a neighbor with probability $64/k$ and a random unpaired image otherwise.
Each micro-batch holds four pairs, eight images, and mixes all 28 pairs of its images.
Four of these mixtures therefore combine two paired, similar images, and the other 24 combine images from different pairs.

\subsection{Interpolation Coefficients}
\label{app:method:alpha}

Each of the 28 mixtures draws one $\alpha \sim \mathrm{Beta}(2,2)$, with density $6\alpha(1-\alpha)$, which favors the middle of the line, where decoding is hardest, Fig.~\ref{fig:analysis}a. \citet{xu26ifid} also evaluate \ifid at the midpoint between two codes, on a spherical path.

\subsection{Evaluation Protocol}
\label{app:results:anchors}

\textbf{Evaluators and samples.}
Every \ifid and ImageNet \rfid we measure in Tables~\ref{tab:main} and~\ref{tab:recon} uses the released evaluation code of \citet{xu26ifid} at its defaults, and daggered values are quoted, as each caption states.
Table~\ref{tab:protocol} gives the corpus, count, pairing and reference of the main measurements, and Table~\ref{tab:sources} the evaluators.

\input{tables/B7_protocol.tex}
\input{tables/B1_anchors.tex}

\textbf{Reproduction of released values.}
Table~\ref{tab:anchors} reproduces the released \ifid of \maetok and DM-VAE and the released \rfid to within a few percent. These are reproduction gaps, not variance intervals.
\citet{chen25maetok} report an rFID of 0.48 for these weights, under a protocol that names neither its reference statistics nor its toolkit.

\begin{table}[!htb]
\caption{\textbf{Evaluators and provenance of the quoted tokenizers.} The evaluator of every measurement, and for each quoted tokenizer what supervised it, on which corpus and at how many code dimensions.}
\label{tab:sources}
\centering\scriptsize\setlength{\tabcolsep}{3pt}
\begin{tabular}{@{}p{0.155\linewidth}|p{0.825\linewidth}@{}}
\toprule
\textbf{Item} & \textbf{Value} \tabularnewline
\midrule
\multicolumn{2}{@{}l}{\textit{Evaluators}} \tabularnewline
\raggedright \ifid\ and \rfid & \raggedright the released evaluation code of \citet{xu26ifid} at its defaults \tabularnewline
\raggedright Inception features & \raggedright the Inception weights of pytorch-fid and of torch-fidelity \tabularnewline
\raggedright Other FID comparisons & \raggedright the ADM evaluator, against the validation originals for reconstruction and interpolation and the ADM ImageNet 256 reference statistics for generation \tabularnewline
\raggedright Pixel metrics & \raggedright our own pixel evaluation, SSIM with a uniform window at data range 2, \lpips\ with AlexNet \tabularnewline
\midrule
\multicolumn{2}{@{}l}{\textit{Encoder supervision, corpus and code dimensions}} \tabularnewline
\raggedright RAE & \raggedright frozen DINOv2 encoder with a trained decoder, ImageNet, 196{,}608 dimensions \tabularnewline
\raggedright DM-VAE & \raggedright DINOv2 encoder with a trained projector and decoder and a distribution-matching objective, ImageNet \tabularnewline
\raggedright \maetok & \raggedright encoder trained on ImageNet against frozen DINOv2 and SigLIP features, 4{,}096 dimensions \tabularnewline
\raggedright VA-VAE, SoftVQ-VAE & \raggedright encoder trained on ImageNet against frozen DINOv2 features, with MAE and CLIP reported as alternatives \tabularnewline
\raggedright l-DeTok & \raggedright no foundation model, ImageNet \tabularnewline
\raggedright SD-VAE, EQ-VAE & \raggedright no foundation model, OpenImages, EQ-VAE an equivariant fine-tune of SD-VAE. Table~\ref{tab:main} quotes SD-VAE \ifid\ for the Stable Diffusion 1.5 autoencoder, while our own SD-VAE rows use its released EMA fine-tune \tabularnewline
\raggedright REPA-E & \raggedright SD-VAE tuned end to end with a diffusion transformer under representation alignment, ImageNet, from SD-VAE's OpenImages weights \tabularnewline
\raggedright \ours & \raggedright initialized from \maetok, whose encoder was trained against DINOv2-L and SigLIP features, the frozen DINOv3-B/16 vision model, a frozen DINO ViT-S/16 discriminator backbone and a representation Fr\'echet loss across three feature spaces, ImageNet, 4{,}096 dimensions \tabularnewline
\midrule
\multicolumn{2}{@{}l}{\textit{Rows of the released comparison omitted from Table~\ref{tab:main}, with their released \ifid}} \tabularnewline
\raggedright QW-VAE & \raggedright 30.58 \tabularnewline
\raggedright SD3-VAE & \raggedright 37.13 \tabularnewline
\raggedright VA-VAE with 64 channels & \raggedright 37.14 \tabularnewline
\raggedright IN-VAE & \raggedright 41.06 \tabularnewline
\raggedright FLUX-VAE & \raggedright 67.41 \tabularnewline
\bottomrule
\end{tabular}
\end{table}

\textbf{Quoted comparisons.}
\label{app:fair}
Table~\ref{tab:sources} names the quoted tokenizers, what supervised each, its corpus and its code dimensions, and every comparison we measure holds the corpus and the evaluator fixed within a column, while daggered values follow the protocol of their source.
The daggered \ifid values of Table~\ref{tab:main} come from the comparison table released with the code of \citet{xu26ifid}, whose values for \maetok and DM-VAE their released code reproduces, Table~\ref{tab:anchors}. It lists thirteen tokenizers in fourteen rows, since VA-VAE appears twice, with codes of 32 and of 64 channels.
Table~\ref{tab:main} omits the 64-channel VA-VAE~\citep{yao25vavae} and the reconstruction-only QW-VAE, SD3-VAE, IN-VAE and FLUX-VAE, whose values Table~\ref{tab:sources} lists.

\textbf{Pixel evaluation.}
Table~\ref{tab:recon} uses our own pixel evaluation on the same ImageNet images, since the released code maps images to $[0,1]$ before treating them as $[-1,1]$.
It follows \maetok's SSIM convention, a uniform window at data range 2, and computes \lpips on AlexNet. Table~\ref{tab:evalvar} gives the other conventions.

\begin{table}[!htb]
\caption{\textbf{Other pixel conventions and other pairings.} ImageNet validation at $n = 50{,}000$ and \ours\ (82k) unless a row states otherwise.}
\label{tab:evalvar}
\centering\scriptsize\setlength{\tabcolsep}{4pt}
\begin{tabular}{@{}p{0.52\linewidth}rrr@{}}
\toprule
Other pixel conventions & \maetok & \ours & SD-VAE \tabularnewline
\midrule
\raggedright SSIM, Gaussian window at data range 1, ImageNet & 0.644 & 0.670 & 0.708 \tabularnewline
\raggedright SSIM, Gaussian window at data range 1, MS-COCO train2017 & 0.656 & 0.682 & 0.724 \tabularnewline
\bottomrule
\end{tabular}
\par\medskip
\begin{tabular}{@{}p{0.52\linewidth}rr@{}}
\toprule
Measurement & \maetok & \ours \tabularnewline
\midrule
\raggedright \lpips\ with VGG, ImageNet & 0.191 & 0.172 \tabularnewline
\raggedright multi-scale SSIM, ImageNet, \ours\ (41k) & 0.929 & 0.934 \tabularnewline
\midrule
\multicolumn{3}{@{}l}{\textit{FID of decoded midpoints under other pairings}} \tabularnewline
\raggedright each model's nearest validation neighbor & 17.28 & 5.82 \tabularnewline
\raggedright index-adjacent pairing under the dataset's class order & 35.12 & 4.34 \tabularnewline
\raggedright random cross-class pairs of Table~\ref{tab:kway} & 87.49 & 14.82 \tabularnewline
\raggedright random cross-class pairs of Table~\ref{tab:appB_ladder} & 87.34 & 14.92 \tabularnewline
\bottomrule
\end{tabular}
\end{table}

\textbf{Pairing.}
The released \ifid evaluation pairs each validation code with one of its ten nearest training codes, chosen with one seed for every model.
It decodes the midpoint of each pair along a spherical path and computes the FID of the decoded midpoints.
Training mixes codes along the straight line of Eq.~\ref{eq:lerp}.
Our own comparison fixes the corpus, the reference, the ADM evaluator~\citep{dhariwal21adm} and the straight line and varies only the pairing, Table~\ref{tab:evalvar}.
The encoders that Table~\ref{tab:main} sets apart are DINOv2 models pretrained on a dataset far larger than ImageNet.
Neither \maetok nor \ours was trained on MS-COCO.

\textbf{Generator training and evaluation.}
Table~\ref{tab:config:sit} gives the SiT-B settings, and \maetok's authors report no SiT-B.
Two draws of one generator differ by up to 0.3 \gfid, and a dash marks a value that is not measured.

\FloatBarrier

%% file: tables/A1_config.tex
\begin{table}[!htb]\noindent\begin{minipage}{\linewidth}\centering
\captionsetup{type=table}
\caption{\textbf{Tokenizer training configuration.} The ImageNet fine-tune behind every reported \ours result.}
\label{tab:config}
\centering\scriptsize\setlength{\tabcolsep}{3pt}
\begin{tabular}{@{}p{0.185\linewidth}|p{0.795\linewidth}@{}}
\toprule
\textbf{Configuration} & \textbf{Value} \tabularnewline
\midrule
\multicolumn{2}{@{}l}{\textit{Data and pairing}} \tabularnewline
\raggedright Training images & \raggedright ImageNet-1k train at $256\times256$ in $[-1,1]$ \tabularnewline
\raggedright Pair sampling & \raggedright each epoch, every image is paired with its most similar unpaired image among its $k$ nearest neighbors, $k=8,\,8,\,16,\,16,\,32,\,32,\,64,\,64,\,128,\,128$ over epochs 1--10 \tabularnewline
\raggedright $\alpha$ sampler & \raggedright Beta(2,2), one $\alpha$ per mixture \tabularnewline
\midrule
\multicolumn{2}{@{}l}{\textit{Tokenizer and batches}} \tabularnewline
\raggedright Base model & \raggedright \maetok, encoder and decoder trainable \tabularnewline
\raggedright Latent & \raggedright 128 tokens $\times$ 32 channels, unconstrained (no KL, no quantization) \tabularnewline
\raggedright Appended decoder blocks & \raggedright 12 $\to$ 18 blocks, copies of blocks 6--11, each behind a gate initialized at zero, so that the decoder starts identical to \maetok's, $+43.1$M parameters \tabularnewline
\raggedright Micro-batch & \raggedright 8 images and their 28 pairs, so 8 reconstructions and 28 mixtures decoded together \tabularnewline
\raggedright Global batch & \raggedright 256 images per optimizer step \tabularnewline
\midrule
\multicolumn{2}{@{}l}{\textit{Optimization}} \tabularnewline
\raggedright Tokenizer & \raggedright AdamW, lr $10^{-5}$, $\beta=(0.9,0.99)$, weight decay 0.01, constant schedule, grad clip 1.0 \tabularnewline
\raggedright Discriminator & \raggedright AdamW, lr $10^{-4}$, $\beta=(0.9,0.95)$, weight decay 0, grad clip 1.0 \tabularnewline
\raggedright Weight average & \raggedright exponential moving average at decay 0.998, updated every optimizer step, and every \ours number in this paper is measured on the averaged weights \tabularnewline
\raggedright Seed & \raggedright 0 \tabularnewline
\midrule
\multicolumn{2}{@{}l}{\textit{Compute}} \tabularnewline
\raggedright Hardware & \raggedright 8 NVIDIA A100 GPUs with 40\,GB memory each \tabularnewline
\raggedright Training time & \raggedright 72\,h for 41{,}000 steps \tabularnewline
\raggedright Resumed training & \raggedright resumed from step 41{,}000, a further 72\,h to step 82{,}000 \tabularnewline
\bottomrule
\end{tabular}
\end{minipage}\end{table}

\begin{table}[!htb]\noindent\begin{minipage}{\linewidth}\centering
\captionsetup{type=table}
\caption{\textbf{Discriminator, vision model and generator settings.} The frozen backbones act during training only.}
\label{tab:config:critic}\label{tab:config:sit}
\centering\scriptsize\setlength{\tabcolsep}{3pt}
\begin{tabular}{@{}p{0.185\linewidth}|p{0.795\linewidth}@{}}
\toprule
\textbf{Configuration} & \textbf{Value} \tabularnewline
\midrule
\multicolumn{2}{@{}l}{\textit{Discriminator}} \tabularnewline
\raggedright Discriminator backbone & \raggedright frozen DINO ViT-S/16 at 224\,px, features from 5 layers, 196 tokens per layer \tabularnewline
\raggedright Discriminator heads & \raggedright the interpolation discriminator's head, which scores the decoded mixture beside one source, and the unconditional reconstruction discriminator's head, both on the shared backbone \tabularnewline
\raggedright Discriminator parameters & \raggedright 25.9M trainable, frozen backbone 21.7M \tabularnewline
\raggedright Adversarial schedule & \raggedright the discriminator trains from step 0 and its loss reaches the tokenizer after 1{,}500 steps, one discriminator update per tokenizer update, adaptive weight capped at 1.0 \tabularnewline
\raggedright Adaptive weight & \raggedright the VQGAN ratio $\|\nabla \mathcal{L}_{\mathrm{rec}}\| / (\|\nabla \mathcal{L}_G\| + 10^{-4})$ at the decoder's final linear layer, capped at 1.0 \tabularnewline
\raggedright Discriminator regularizers & \raggedright LeCAM $10^{-3}$ at EMA decay 0.999, consistency regularization at weight 4.0, not scaled by the adversarial weight 0.4, DiffAug color, translation and cutout at probability 0.5 on both networks, and no gradient penalty \tabularnewline
\raggedright Head per layer & \raggedright spectral-normalized~\citep{miyato18sn} 1-d convolutions, a $k{=}1$ projection, local batch normalization and LeakyReLU at slope 0.2, a residual $k{=}9$ circular convolution block scaled by $1/\sqrt2$, and a $k{=}1$ projection to one logit per token \tabularnewline
\raggedright Source cross-attention & \raggedright pre-layer-normalized spectral-normalized linear maps, 6 heads, scaled dot-product attention, a position-wise MLP at $4\times$ expansion with GELU, LayerScale initialized at $10^{-4}$ on both residual branches so the block is the identity at initialization, and a $k{=}1$ spectral-normalized convolution fusing $[x;\ \mathrm{att};\ x - \mathrm{att}]$ \tabularnewline
\raggedright Passes per mixture & \raggedright four, two in which a source attends to the mixture and two in which the mixture attends to a source, with the two source scores and the two mixture scores averaged before the hinge loss, and all four passing gradient to the tokenizer \tabularnewline
\midrule
\multicolumn{2}{@{}l}{\textit{Representation Fr\'echet loss}} \tabularnewline
\raggedright Fr\'echet moments & \raggedright moving average of the feature moments of the decoded images at decay 0.999, blended with the current batch \tabularnewline
\raggedright Fr\'echet warm-up & \raggedright the moments accumulate for 500 optimizer steps before the term enters the loss \tabularnewline
\midrule
\multicolumn{2}{@{}l}{\textit{Frozen vision model}} \tabularnewline
\raggedright Backbone & \raggedright frozen DINOv3 ViT-B/16 pretrained on LVD-1689M, CLS feature precomputed for every training image, patch features computed during training \tabularnewline
\raggedright Losses & \raggedright distance-matrix alignment 0.5, latent-CLS alignment 0.25 and decoder-patch alignment 0.1 on single images, the feature mixing ratio loss 0.5 on at most 8 of the 28 mixtures of a micro-batch \tabularnewline
\midrule
\multicolumn{2}{@{}l}{\textit{Generator}} \tabularnewline
\raggedright Model & \raggedright SiT-B/1D~\citep{ma24sit} on each tokenizer's codes, one generator per tokenizer \tabularnewline
\raggedright Training steps & \raggedright 400{,}000 steps with the training code of \citet{xu26ifid} \tabularnewline
\raggedright Seed & \raggedright seed 0, one training run per tokenizer, and seed 1 for the second generator of App.~\ref{app:results:seeds} \tabularnewline
\midrule
\multicolumn{2}{@{}l}{\textit{Sampling and evaluation}} \tabularnewline
\raggedright Sampler & \raggedright stochastic sampler of SiT at 250 steps, classifier-free guidance scale 1.0 (no guidance) \tabularnewline
\raggedright Samples & \raggedright 50{,}000 class-conditional samples per tokenizer \tabularnewline
\raggedright Metric & \raggedright FID-50k against the ADM reference statistics for ImageNet $256\times256$, compared with our own \maetok\ control, App.~\ref{app:results:gfid} \tabularnewline
\bottomrule
\end{tabular}
\end{minipage}\end{table}

%% file: tables/B7_protocol.tex
\begin{table}[!htb]
\caption{\textbf{Evaluation datasets and protocols.} One row per measurement, with its corpus, sample count, pairing and reference.}
\label{tab:protocol}
\centering\scriptsize\setlength{\tabcolsep}{3pt}
\begin{tabular}{@{}p{0.115\linewidth}p{0.125\linewidth}p{0.395\linewidth}p{0.315\linewidth}@{}}
\toprule
Split & Units & Source and pairing & Evaluation target \tabularnewline
\midrule
\raggedright ImageNet validation & \raggedright 50{,}000 midpoints & \raggedright val code to a sampled top-10 nearest neighbor in the train split, spherical path, $\alpha=0.5$, the released evaluation code at its defaults, FID against the 50{,}000 validation originals & \raggedright \ifid \tabularnewline
\raggedright ImageNet validation & \raggedright 50{,}000 images & \raggedright reconstruction, no pairing, the released evaluation code at its defaults for \rfid\ and our own pixel evaluation, SSIM with a uniform window at data range 2, for the rest & \raggedright \rfid\ (Tables~\ref{tab:main} and~\ref{tab:recon}), PSNR, SSIM, \lpips\ (Table~\ref{tab:recon}) \tabularnewline
\raggedright ImageNet validation & \raggedright 50{,}000 reference images & \raggedright equal-weight barycenters of $k=2,3,5,10$ codes, same-class and cross-class, straight line & \raggedright FID, sFID, precision and recall of decoded barycenters (Table~\ref{tab:kway}) \tabularnewline
\raggedright ImageNet validation & \raggedright 50{,}000 midpoints & \raggedright random cross-class pairs under one seed, straight-line midpoint at $\alpha=0.5$, ADM evaluator against the 50{,}000 validation originals & \raggedright FID of decoded midpoints through the fine-tune (Table~\ref{tab:appB_ladder}) \tabularnewline
\raggedright ImageNet validation & \raggedright 5{,}000 near and 5{,}000 random pairs & \raggedright near pairs at DINOv3 cosine 0.6 to 0.8 and random pairs of different classes under one seed, the straight line decoded at nine equally spaced positions, \lpips\ with AlexNet between consecutive decodes and to both source reconstructions & \raggedright perceptual path length, largest-step share, midpoint distance and ordered fraction (Table~\ref{tab:ppl}) \tabularnewline
\raggedright MS-COCO train2017 & \raggedright 50{,}000 images & \raggedright reconstruction, no pairing & \raggedright \rfid, PSNR, SSIM, \lpips\ (Table~\ref{tab:recon}) \tabularnewline
\raggedright MS-COCO train2017 & \raggedright 50{,}000 midpoints & \raggedright one random pairing of different images shared by every model, straight-line midpoint at $\alpha=0.5$ & \raggedright FID of decoded midpoints, and the same ranking in four feature spaces (App.~\ref{app:results:ood}) \tabularnewline
\raggedright MS-COCO val2017 & \raggedright 5{,}000 images & \raggedright reconstruction, no pairing, the split \maetok\ reports on & \raggedright \rfid\ against \maetok's published COCO value \tabularnewline
\raggedright ImageNet class-conditional & \raggedright 50{,}000 samples & \raggedright SiT-B/1D, 400k steps, 250 steps of the stochastic sampler, no guidance & \raggedright \gfid\ (App.~\ref{app:results:gfid}) \tabularnewline
\raggedright ImageNet validation & \raggedright 50{,}000 images & \raggedright MAE linear probe, 90 epochs, LARS & \raggedright top-1 accuracy (Table~\ref{tab:appB_ladder}) \tabularnewline
\bottomrule
\end{tabular}
\end{table}

%% file: tables/B1_anchors.tex
\begin{table}[!htb]
\caption{\textbf{Reproducing published measurements.} Each row re-runs one published measurement under the protocol that produced it. $^{\dagger}$The published value.}
\label{tab:anchors}
\centering\scriptsize\setlength{\tabcolsep}{4pt}
\begin{tabular}{@{}p{0.29\linewidth}rrrp{0.20\linewidth}@{}}
\toprule
Quantity & reported & ours & rel.\ diff.\ & source \tabularnewline
\midrule
\raggedright \ifid, \maetok & 14.06$^{\dagger}$ & 14.42 & $+$2.6\% & \raggedright comparison table released by \citet{xu26ifid} \tabularnewline
\raggedright \ifid, DM-VAE & 8.14$^{\dagger}$ & 8.29 & $+$1.8\% & \raggedright comparison table released by \citet{xu26ifid} \tabularnewline
\raggedright \rfid, \maetok, \ifid\ protocol & 0.58$^{\dagger}$ & 0.5819 & $+$0.3\% & \raggedright released leaderboard of \citet{xu26ifid} \tabularnewline
\raggedright \rfid, \maetok, MS-COCO & 4.87$^{\dagger}$ & 5.05 & $+$3.8\% & \raggedright Table 4 of \citet{chen25maetok} \tabularnewline
\bottomrule
\end{tabular}
\end{table}